%% file: main.tex
\documentclass[10pt,twocolumn,letterpaper]{article}
\usepackage{amssymb}
\usepackage{pifont}
\usepackage[table,xcdraw,dvipsnames]{xcolor} 
\usepackage{colortbl}
\usepackage[pagenumbers]{cvpr} 
\usepackage{multirow}
\definecolor{cvprblue}{rgb}{0.21,0.49,0.74}
\usepackage[pagebackref,breaklinks,colorlinks,allcolors=cvprblue]{hyperref}

\title{MeshCraft: Exploring Efficient and Controllable Mesh Generation with Flow-based DiTs}

\author{
    Xianglong He$^{1}$ \qquad
    Junyi Chen$^{2, 3}$ \qquad
    Di Huang$^{4}$ \qquad
    Zexiang Liu$^{5}$ \qquad
    Xiaoshui Huang$^{2}$ \\
    Wanli Ouyang$^{6}$ \qquad
    Chun Yuan$^{1\dag}$ \qquad
    Yangguang Li$^{5\dag}$ \\
    $^1${Tsinghua University} \qquad $^2${Shanghai Jiaotong University} \qquad $^3${Shanghai AI Laboratory} \\ $^4${The University of Sydney} \qquad $^5${VAST} \qquad
    $^6${The Chinese University of Hong Kong} \\
}

\begin{document}

\twocolumn[{%
\renewcommand\twocolumn[1][]{#1}%
\maketitle
\begin{center}
    \vspace{-0.7cm}
    \centering
    \captionsetup{type=figure}
    \includegraphics[width=\textwidth]{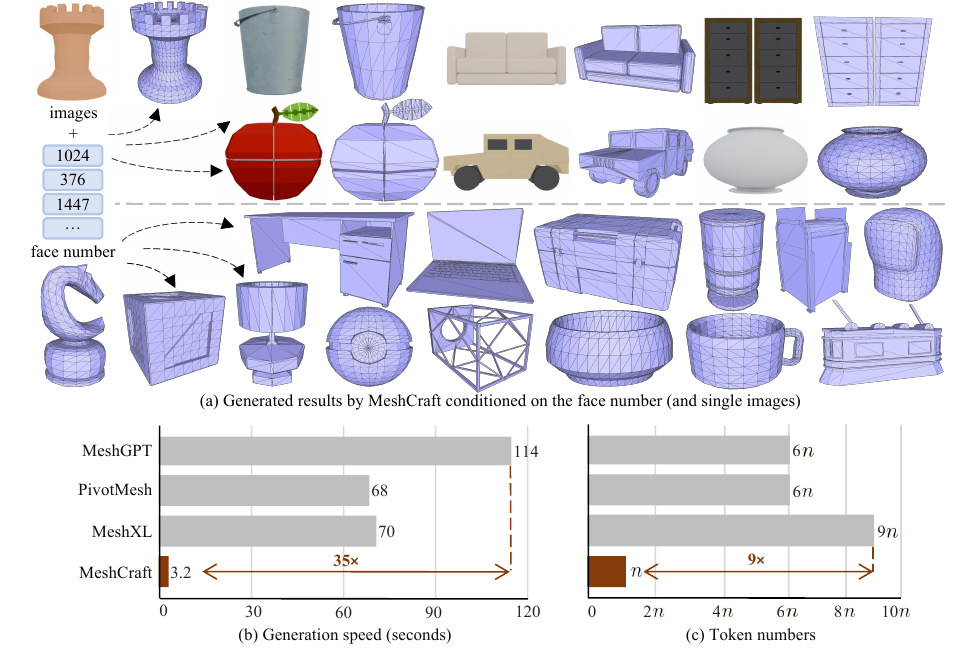}
    \vspace{-0.8cm}
    \captionof{figure}{$\mathbf{Overview}$ of generated meshes, speed and token numbers of MeshCraft.}
    \vspace{-0.2cm}
\label{fig:teaser}
\end{center}%
}]
\def\thefootnote{}\footnotetext{$\dag$ Corresponding authors.} 
\input{sec/0_abstract} 
\input{sec/1_introduction}

\input{sec/2_related_works}

\input{sec/3_preliminary}
\input{sec/4_methodology}
\input{sec/5_experiments}

\input{sec/6_conclusions}
{
    \small
    \bibliographystyle{ieeenat_fullname}
    \bibliography{main}
}
\clearpage
\input{sec/X_suppl}

\end{document}

%% file: sec/0_abstract.tex
\begin{abstract}
In the domain of 3D content creation, achieving optimal mesh topology through AI models has long been a pursuit for 3D artists. 
Previous methods, such as MeshGPT, have explored the generation of ready-to-use 3D objects via mesh auto-regressive techniques. 
While these methods produce visually impressive results, their reliance on token-by-token predictions in the auto-regressive process leads to several significant limitations. These include extremely slow generation speeds and an uncontrollable number of mesh faces.
In this paper, we introduce MeshCraft, a novel framework for efficient and controllable mesh generation, which leverages continuous spatial diffusion to generate discrete triangle faces.
Specifically, MeshCraft consists of two core components: 1) a transformer-based VAE that encodes raw meshes into continuous face-level tokens and decodes them back to the original meshes, and 2) a flow-based diffusion transformer conditioned on the number of faces, enabling the generation of high-quality 3D meshes with a predefined number of faces.
By utilizing the diffusion model for the simultaneous generation of the entire mesh topology, MeshCraft achieves high-fidelity mesh generation at significantly faster speeds compared to auto-regressive methods. Specifically, MeshCraft can generate an 800-face mesh in just 3.2 seconds—35$\times$ faster than existing baselines. Extensive experiments demonstrate that MeshCraft outperforms state-of-the-art techniques in both qualitative and quantitative evaluations on ShapeNet dataset and demonstrates superior performance on Objaverse dataset. Moreover, it integrates seamlessly with existing conditional guidance strategies, showcasing its potential to relieve artists from the time-consuming manual work involved in mesh creation.
\end{abstract}

%% file: sec/1_introduction.tex
\section{Introduction}
\label{sec:intro}

With advances in fields such as gaming and 3D printing, the creation of high-quality, topologically sound 3D meshes has become increasingly important. However, generating well-structured 3D meshes from simple inputs—such as textual descriptions or a single 2D image—requires significant artistic expertise and often entails labor-intensive manual processes. Therefore, the ability to rapidly generate 3D meshes with high-quality topology that meet the needs of artists and professionals using AI models is a key goal.

Significant efforts have been made to automate the generation of 3D meshes. Many works~\cite{poole2022dreamfusion, xu2024grm} are modeling with dense triangle meshes, which are extracted from neural fields using iso-surfacing methods \cite{mildenhall2021nerf, park2019deepsdf, kerbl20233d}. Specifically, these methods first generate intermediate 3D representations and then post-process them into final meshes. Although these approaches yield impressive visualizations in producing neural representations, the resulting meshes often suffer from excessive face counts and artifacts due to misalignment that occur during the conversion of intermediate representations into final meshes via re-meshing techniques~\cite{lorensen1998marching}. The efficiency of these processes is also limited by the post-processing step.

To address these issues, another group of works~\cite{nash2020polygen, siddiqui2024meshgpt, alliegro2023polydiff, hao2024meshtron}, modeling with triangle meshes that accurately reflect the compactness of artists' design, offers great flexibility for manipulation and efficiency for storage. This approach, referred to as "native mesh generation," focuses on explicitly modeling mesh distributions. It first transforms the mesh into latent sequences by leveraging the spatial relationships among faces, vertices, and coordinates. These methods then directly generate the sequences, and even shows superior long sequence scalability. However, these approaches are often limited by generation effectiveness or are validated only on small datasets, such as ShapeNet~\cite{chang2015shapenet}. Furthermore, most of the methods lack user controllability, which is essential for the 3D creation process.

To address these challenges, we introduce MeshCraft, an efficient and controllable approach for high-fidelity mesh generation method that leverages the strengths of continuous diffusion transformers with rectified flow (see \cref{fig:teaser}(b) and (c) for comparisons with prior works). Unlike auto-regressive methods such as Meshtron~\cite{hao2024meshtron}, which prioritize scalability, we explore the possibilities of different routes starting from the underlying principles of implementing native mesh generation, with the goal of demonstrating the potential of diffusion models for the practical usage of controllable and efficient mesh generation, rather than immediate scalability.

MeshCraft saves token numbers for at most 9 times, and speed up for 35 times. It is designed as a two-stage pipeline that consists of directly modeling the mesh distribution in a latent space via a Variational Auto-Encoder(VAE)~\cite{kingma2013auto} and then generating meshes with a diffusion transformer. Unlike previous works that compress meshes into discrete tokens, we explore the modeling of meshes in a low-dimensional and continuous latent space. Face features of the mesh are sent into the VAE encoder, regularized with KL divergence, and are subsequently decoded into vertex coordinates of each face. For the generative model, the diffusion transformer is modified to generate the continuous tokens with varying lengths with inspirations from recent advances in image generation~\cite{esser2024scaling, lu2024fit}.
This approach enables MeshCraft to generate meshes at high speed. The guidance and masking strategy also allow users to control the number of faces of generated meshes in a user-friendly manner.

We validate the effectiveness of MeshCraft on the ShapeNet dataset~\cite{chang2015shapenet} and demonstrate its great potential as a generic mesh generator on Objaverse~\cite{deitke2023objaverse}, which includes a diverse array of 3D objects across various categories. Extensive experiments illustrate the effectiveness, and controllability of MeshCraft from both qualitative and quantitative aspects.
In summary, our contributions are as follows:

\begin{itemize}
\item We propose a transformer-based VAE that encodes discrete triangle meshes into a continuous latent space and decodes them back to the original mesh, achieving competitive reconstruction performance compared to state-of-the-art mesh auto-encoders based on vector quantization.

\item We introduce a flow-based transformer diffusion model conditioned on the number of faces, integrating classifier-free guidance for both image inputs and face number to enable effective control over the mesh generation process.

\item MeshCraft significantly outperforms existing methods, achieving new state-of-the-art results on the ShapeNet dataset while being 35$\times$ faster than MeshGPT, and demonstrating the effectiveness of face number control on the large-scale Objaverse dataset.

\end{itemize}

%% file: sec/2_related_works.tex
\section{Related works}
\label{sec:related_works}
Recent advances~\cite{nash2020polygen, alliegro2023polydiff, siddiqui2024meshgpt, weng2024pivotmesh, chen2024meshxl, hao2024meshtron} have seen pioneering efforts to generate meshes directly, with many employing auto-regressive models for this task. For instance, PolyGen~\cite{nash2020polygen} utilizes two transformers to separately learn vertex and face distributions. MeshGPT~\cite{siddiqui2024meshgpt} first encodes meshes into face-level quantized tokens via GNN-based Vector Quantization Variational Auto-Encoder(VQ-VAE)~\cite{van2017neural}, followed by the application of GPT-style transformers for auto-regressive generation. MeshXL~\cite{chen2024meshxl} introduces another sequential mesh representation for one-stage auto-regressive generation. Even though Meshtron~\cite{hao2024meshtron} shows the scalability of auto-regressive generations, however, the generative capabilities of these models are restricted by the rapidly increasing number of tokens, leading to slow inference speeds. Additionally, users cannot precisely assign an exact number of object faces, which limits their practicality in the existing workflow for 3D creations.
Among these works, Polydiff~\cite{alliegro2023polydiff} is the most relevant to our approach. It trains a class-conditioned discrete diffusion model using discrete state transition matrices and cross-entropy loss, which is challenging to optimize and hard to benefit from existing conditional guidance method. In contrast, we decouple modeling and generation: first compressing meshes into a semantically rich latent space with a KL-regularized VAE, then applying a flow-based DiT to model this space for mesh generation. MeshCraft is the first to introduce a fine-grained controllable mesh generator based on diffusion transformers.

\subsection{Image-to-3D Generation}
With the effective exploration of 3D representations~\cite{mildenhall2021nerf, park2019deepsdf, kerbl20233d}, large-scale datasets~\cite{deitke2023objaverse}, and popular 2D generative models~\cite{rombach2022high}, conditional generation of diverse and high-fidelity 3D assets has emerged as a promising area of research. Most studies\cite{poole2022dreamfusion, tang2025lgm, liu2024pi3d, he2025gvgen} first learn neural 3D representations and then post-process~\cite{lorensen1998marching, chen2024meshanything} them into meshes, which can lead to overly dense results. Some approaches~\cite{poole2022dreamfusion, lin2023magic3d, xu2024grm} utilize pre-trained text-to-image models to optimize targeted meshes based on given conditions, resulting in significant time costs. In this paper, we focus on "directly" generating high-fidelity meshes in face-level representations, effectively balancing quality and efficiency.

\subsection{Flexible Diffusion Models}
The rise of text-to-image diffusion models has prompted the consideration of more customized demands, particularly in generating images with unrestricted resolutions. Recent works~\cite{lu2024fit, zhuo2024lumina, chen2024pixart} have explored this area with flexible training, inference strategies, and model designs. Inspired by these advancements, we propose MeshCraft, a controllable mesh generation pipeline based on flow-based diffusion transformers. Our user-friendly method enables the rapid generation of compact meshes while allowing for extensive manipulation of the results, thereby facilitating practical applications of AI-generated 3D assets in the industry.

%% file: sec/3_preliminary.tex
\section{Preliminary}
\label{sec:pre}
\subsection{Ordered mesh representation}
\label{sec:pre_order}
In this work, we adopt the ordering rule used in previous native mesh generation studies~\cite{nash2020polygen, siddiqui2024meshgpt, weng2024pivotmesh} and consider meshes as ordered sequences consisting of three progressively defined components: face level, vertex level, and coordinate level. Let \(\mathcal{M}\) be a mesh that includes \(n\) faces \(\{\mathbf{f}_{i}\}_{i=1,2,\cdots,n}\), where each face \(\mathbf{f}_{i}\) comprises \(k\) vertices, represented as \(\mathbf{f}_{i}=\{\mathbf{v}_{i}^1,\mathbf{v}_{i}^2,\cdots,\mathbf{v}_{i}^k\}\). Each vertex is defined in the coordinate system as \(\mathbf{v}_{i}^j=(x_i^j,y_i^j,z_i^j)\). Consequently, the mesh \(\mathcal{M}\) can be represented in the following manner (taking \(k=3\) as an example):
\begin{equation}
\begin{split}
  \mathcal{M} &= \{\mathbf{f}_{1},\mathbf{f}_2,\cdots,\mathbf{f}_n\}\\
  &= \{\mathbf{v}_{1}^1,\mathbf{v}_{1}^2,\mathbf{v}_{1}^3,\mathbf{v}_{2}^1,\mathbf{v}_{2}^2,\mathbf{v}_{2}^3,\cdots,\mathbf{v}_{n}^1,\mathbf{v}_{n}^2,\mathbf{v}_{n}^3\}\\
  &= \{x_1^1,y_1^1,z_1^1,x_1^2,y_1^2,z_1^2,\cdots,x_n^3,y_n^3,z_n^3\}
\label{eq:1}
\end{split}
\end{equation}
 For sequence ordering, faces are sorted by vertex indices from lowest to highest, while vertices are sorted by their z-y-x coordinates in the same manner to ensure the uniqueness of the mesh. Unlike prior works using representations at the vertex or coordinate levels to generate meshes, we generate face-level tokens using flow-based diffusion transformers. This approach significantly reduces the number of tokens and enhances the efficiency of mesh generation.

\begin{figure*}[!t]
\begin{center}
\vspace{-0.6cm}
\centerline{\includegraphics[width=1.1\linewidth]{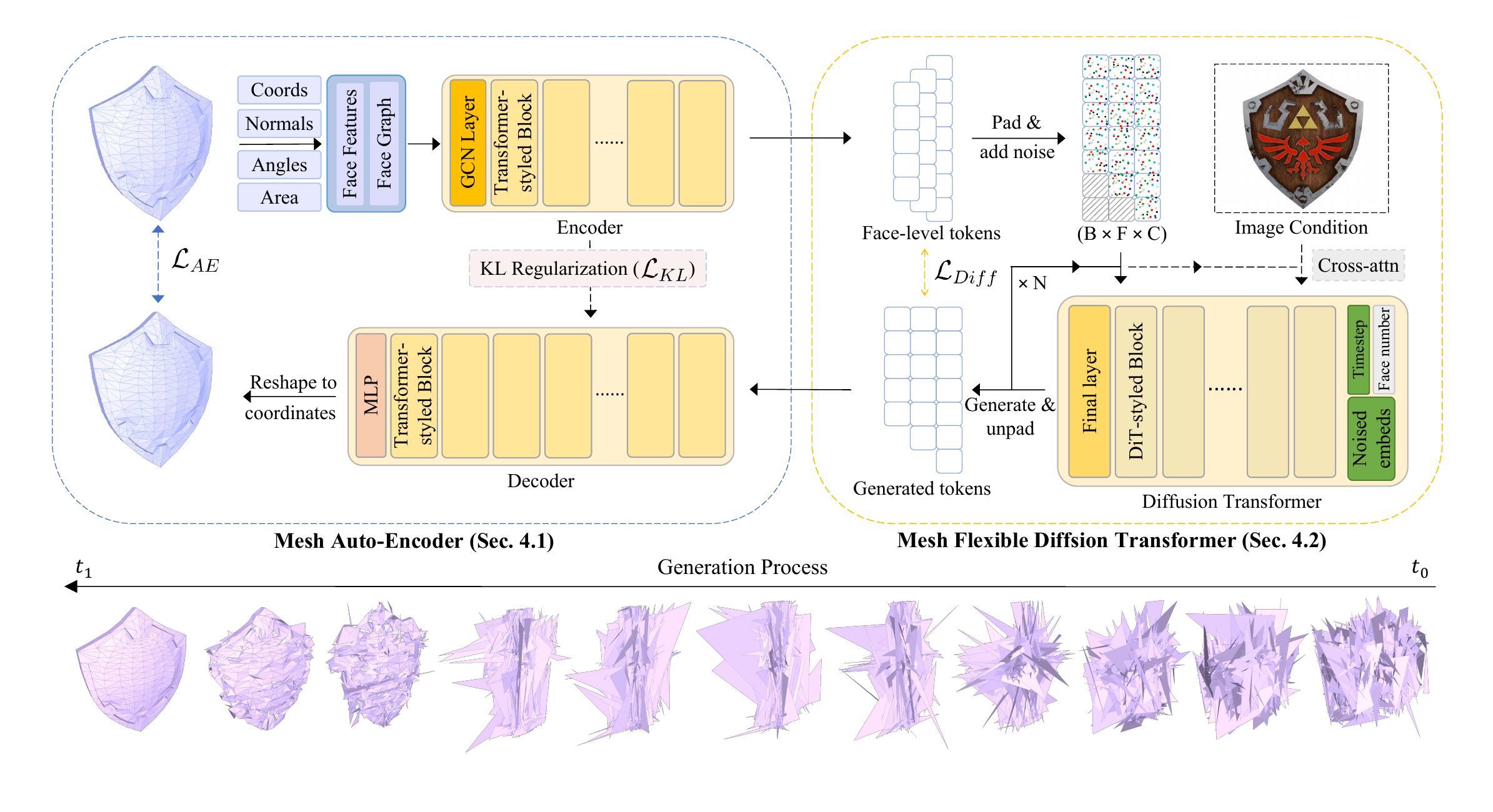}}
\vspace{-1.0cm}
\caption{\textbf{Pipeline of MeshCraft.} Our framework comprises two stages. We firstly compress meshes into face-level tokens (\cref{sec:ae}). Then the tokens are used for training the flow-based DiT, which is guided by the input face number and the image conditions (\cref{sec:dit}).}
\label{fig:pipeline}
\end{center}
\vspace{-1.1cm}
\end{figure*}
\subsection{Rectified flow}
With the growing popularity of generative models in recent years, diffusion models~\cite{rombach2022high, ho2020denoising, song2020score, peebles2023scalable, ma2024sit} have been widely recognized for their powerful modeling capabilities. Score-based models, such as~\cite{song2020score} and DDPM~\cite{ho2020denoising}, are commonly employed, formulating the diffusion process through stochastic differential equations (SDEs). However, these methods often suffer from a slow iterative de-noising process, resulting in inefficient inference time. In contrast, rectified flow~\cite{liu2022flow} is an implicit probabilistic model designed for fast generation, based on ordinary differential equations (ODE). It aims to transport the distribution \(\pi_0\) to \(\pi_1\) by following straight-line paths as much as possible. This preference is both theoretically and computationally advantageous, allowing for few-step or even one-step sampling.

Given two distributions \(\pi_0\) and \(\pi_1\), the rectified flow induced from \((X_0, X_1)\), where \(X_0\sim \pi_0\) and \(X_1\sim \pi_1\), is modeled as an ODE over time \(t\in [0, 1]\),
\begin{equation}
    \mathrm{d} Z_t=v(Z_t,t)\mathrm{d} t
\end{equation}
which converts $Z_0$ from $\pi_0$ to a $Z_1$ following $\pi_1$. The drift force $v: \mathbb{R}^d \rightarrow \mathbb{R}^d$ is to drive the flow to follow the linear direction $X_1-X_0$ as much as possible by solving a simple least squares regression problem:
\begin{equation}
    \min_v \int_0^1{\mathbb{E}\left[ \| (X_1-X_0)-v(X_t,t) \|^2 \right]\mathrm{d} t}
\end{equation}
where $X_t = t X_1 + (1-t) X_0$ is the linear interpolation of $X_0$ and $X_1$. $v$ is parameterized by the models in practice.

Rectified flow not only avoids crossing paths when finding the solution but also reduces errors arising from discrete-time scheduling and transport costs. To leverage rectified flow and advanced model architectures, we follow the implementation in SiT~\cite{ma2024sit} and introduce several techniques to adapt it to the task of mesh generation.

\begin{figure}[!t]
\begin{center}
\centerline{\includegraphics[width=\linewidth]{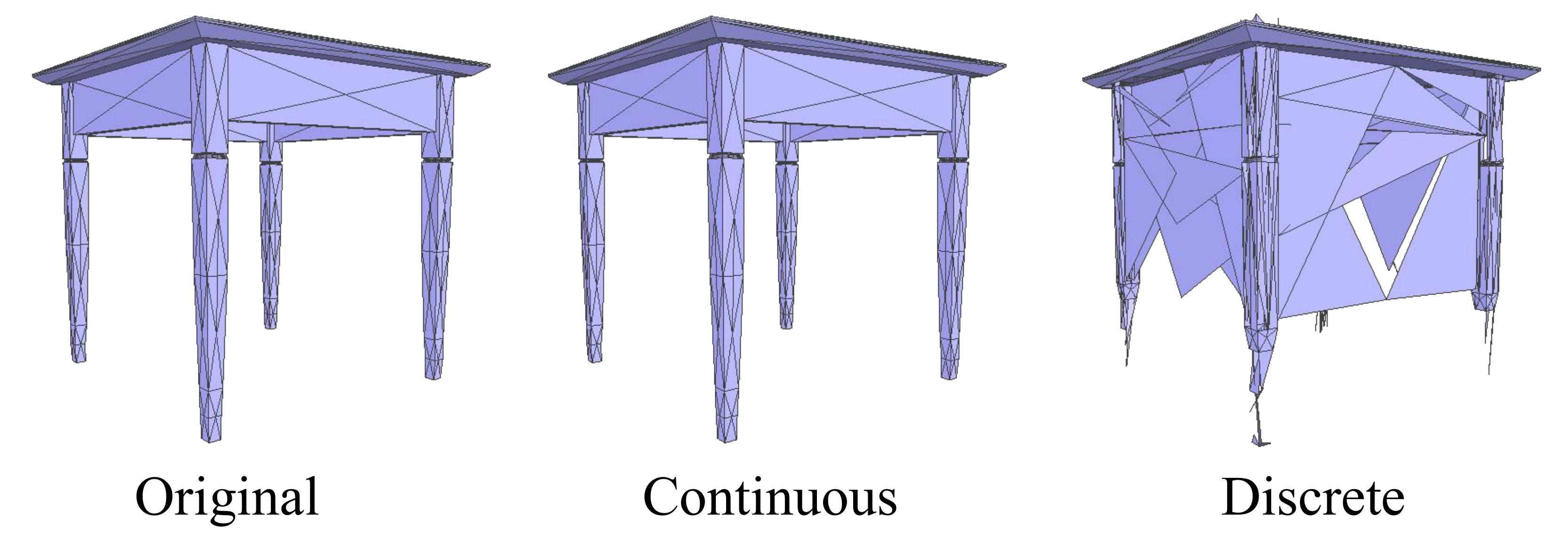}}
\vspace{-0.4cm}
\caption{\textbf{Reconstruction quality using different tokenizers.} "Continuous" means using KL-divergence loss to regularize continuous-space tokens, while "Discrete" stands for using RVQ~\cite{zeghidour2021soundstream} to quantize discrete tokens for reconstruction. Refer to ~\cref{tab:recon} for quantitative results.}
\label{fig:recon}
\end{center}
\vspace{-1.3cm}
\end{figure}

%% file: sec/4_methodology.tex
\section{Methodology}
\label{sec:method}
Existing works~\cite{nash2020polygen, siddiqui2024meshgpt, weng2024pivotmesh, chen2024meshxl} that generate artist-like meshes predominantly utilize auto-regressive models. However, these approaches require discretizing coordinates into a limited vocabulary, resulting in long token sequences for each mesh. This can lead to information loss and inefficient inference speeds.
In contrast, we propose generating latent tokens of faces in the continuous space through a transformer-based auto-encoder, and we model the distribution using flow-based DiTs. As \cref{fig:recon} shows, the continuous tokenizer achieves higher reconstruction accuracy than the discrete one, justifying our choice of continuous diffusion over discrete AR, which is also a foundation motivation for the diffusion-based exploration.
Moreover, this approach facilitates fast and controllable mesh generation. Furthermore, it allows our method to integrate with popular guidance strategies~\cite{ho2022classifier}, enabling control over the diffusion process with different conditions and enhancing generation quality. An overview of MeshCraft is illustrated in \cref{fig:pipeline}.

\begin{figure*}[!t]
\begin{center}
\centerline{\includegraphics[width=1.01\linewidth]{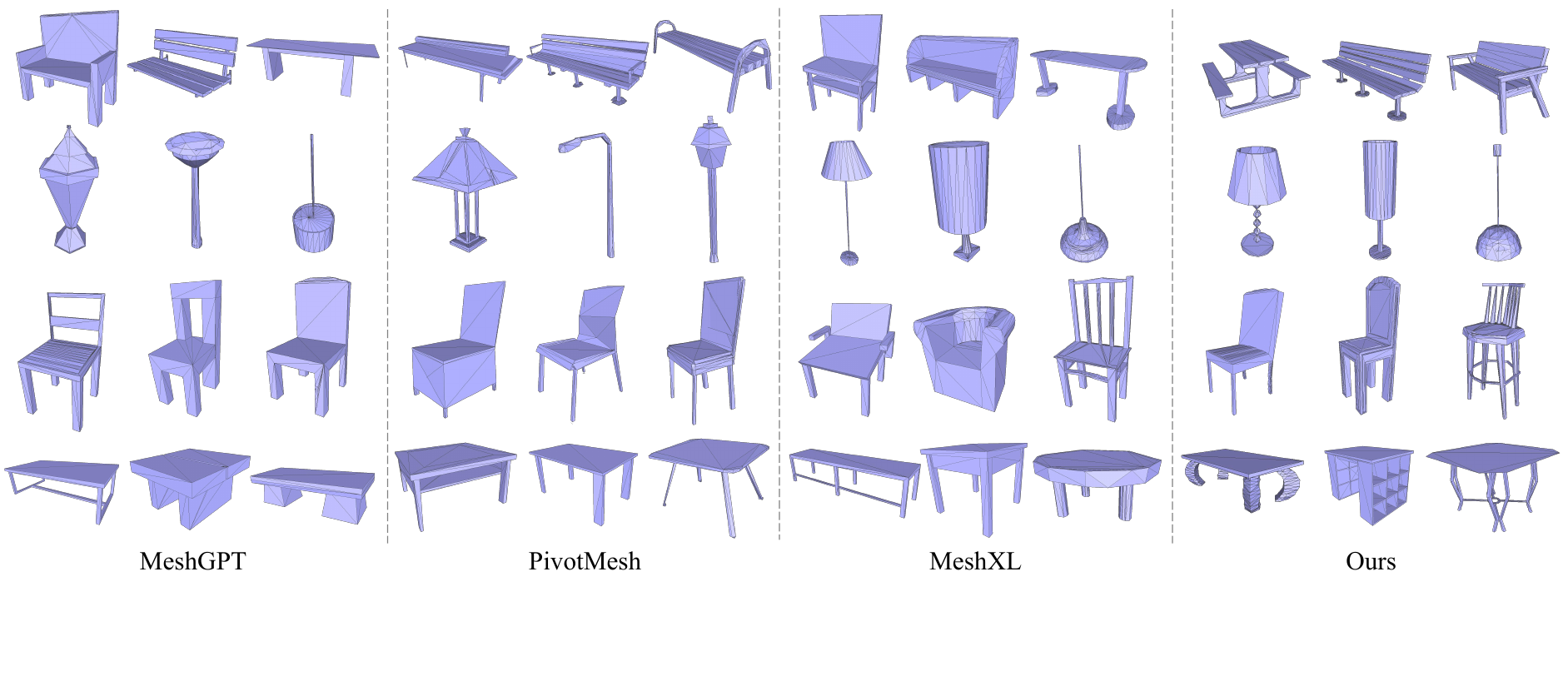}}
\vspace{-2.2cm}
\end{center}
\caption{\textbf{Qualitative comparisons on ShapeNet.} MeshCraft produces high-quality meshes with sharp edges and smooth faces.}
\label{fig:shapenet_qualitative}
\vspace{-0.4cm}
\end{figure*}

\subsection{Encoding meshes into face-level continuous tokens}
\label{sec:ae}
As described in \cref{sec:pre_order}, we formulate meshes as ordered sequences according to \cref{eq:1}. To effectively learn their distribution, the sequences are fed into encoder \(E\) along with associated geometric information (normals, angles, areas, and adjacency among faces), which is aggregated using a single Graph Convolutional Network (GCN) layer, preserving geometric information and enhancing the representation robustness. Subsequently, \(N_E\) transformer-style blocks are employed to extract face-level features \(F_i\).

Unlike previous works~\cite{siddiqui2024meshgpt, weng2024pivotmesh} that utilize residual vector quantization~\cite{zeghidour2021soundstream} to obtain discrete tokens, we linearly project the features into continuous space:
\begin{equation}
\begin{split}
    \mathrm{FC}_{\mu}(F_i) = \left(\mu_{i, j}\right)_{j\in [ 1,2,\cdots, C_{KL}]} \\
    \mathrm{FC}_{\sigma}(F_i) = \left(\log\sigma^2_{i, j}\right)_{j\in [ 1,2,\cdots, C_{KL}]}
\end{split}
\end{equation}
where \(\mathrm{FC}_{\mu}(\cdot)\) and \(\mathrm{FC}_{\sigma}(\cdot)\) are linear projection layers. The continuous tokens \(F_i'\) can then be sampled from \((\mu_i, \sigma_i)\). To better adapt them for training the diffusion model, we use KL-divergence to regularize them:

\begin{equation}
    \mathcal{L}_{KL}\left(\{F_i\}_{i=1}^n\right)=\frac{1}{n\cdot C_{KL}}\sum_{i=1}^{n}\sum_{j=1}^{C_{KL}}\frac{1}{2}\left(\mu^2_{i,j}+\sigma^2_{i,j}-\log \sigma^2_{i,j}\right)
\end{equation}
As the findings of Fluid~\cite{fan2024fluid} claim, this transformation eliminates the need for the codebook, and also results in significantly shorter token lengths, leading to accurate reconstructions (also discussed in \cref{sec:ablation_ae}).

The sampled tokens \(F_i'\) are then passed to the decoder \(D\), which consists of \(N_D\) transformer-style blocks and concludes with an MLP layer. The outputs are reshaped into coordinates to compute the loss \(\mathcal{L}_{AE}\) against the input sequences. Following \cite{siddiqui2024meshgpt}, we use a cross-entropy loss \(\mathcal{L}_{AE}\) to guide the training process.

\begin{table}[t]
\vspace{-0.2cm}
\begin{center}
\begin{tabular}{ccc}
\toprule
\bf Method & \bf Tri. Accu.$(\%)\uparrow$ & \bf L2 Dist.$(\times 10^{-2})\downarrow$ \\
\hline
MeshGPT~\cite{siddiqui2024meshgpt} & \textbf{99.99} & \textbf{0.00} \\
 \hline 
PivotMesh~\cite{weng2024pivotmesh} & 98.88 & 0.86  \\
\hline
\textbf{Ours}    & \underline{99.42} & \underline{0.06} \\
\bottomrule
\end{tabular}
\end{center}
\vspace{-0.6cm}
\caption{\textbf{Reconstruction performance on ShapeNet dataset.} Our continuous auto-encoder behaves competitively with prior works using the vector quantization.}
\vspace{-0.8cm}
\label{tab:recon_shapenet}
\end{table}

\begin{table*}[t]
  \begin{center}
  \vspace{-0.2cm}
    \resizebox{0.95\linewidth}{!}{
      \begin{tabular}{cccccccc|cccccccc} 
        \toprule
        \bf Class & \bf Method & \bf COV$\uparrow$  & \bf MMD$\downarrow$ & \bf 1-NNA & \bf JSD$\downarrow$  & \bf FID$\downarrow$ & \bf KID$\downarrow$ & 
        \bf Class & \bf Method & \bf COV$\uparrow$  & \bf MMD$\downarrow$ & \bf 1-NNA & \bf JSD$\downarrow$  & \bf FID$\downarrow$ & \bf KID$\downarrow$ \\
        \midrule
        \multirow{4}{*}{Chair} & MeshGPT~\cite{siddiqui2024meshgpt} & 45.98 & 10.34 & 60.06 & 11.67  & 25.43  & 4.10  & 
        \multirow{4}{*}{Table} & MeshGPT~\cite{siddiqui2024meshgpt} & 48.85 & 9.23 & \underline{57.82} & \textbf{8.50}  & \underline{21.98}  & 2.99     \\
        & PivotMesh~\cite{weng2024pivotmesh} & 47.99 & \underline{10.00} & 60.06 & 13.51  & 34.40  & 10.33  & 
        & PivotMesh~\cite{weng2024pivotmesh} & 47.42 & \underline{9.08} & 58.35 & 10.42  & 24.97  & 7.99   \\
        & MeshXL*~\cite{chen2024meshxl} & \underline{49.43} & 10.17 & \underline{56.90} & \underline{11.37}  & \textbf{20.09}  & \textbf{1.70}  & 
        & MeshXL*~\cite{chen2024meshxl} & \underline{50.98} & 9.38 & \underline{57.82} & 9.07  & 22.08  & \underline{2.88} \\
        & \textbf{Ours} & \textbf{51.44} & \textbf{9.61} & \textbf{54.31} & \textbf{11.03}  & \underline{20.40}  & \underline{1.76}  & 
        & \textbf{Ours} & \textbf{55.42} & \textbf{8.74} & \textbf{54.26} & \underline{8.73}  & \textbf{16.63}  & \textbf{1.70} \\
        \midrule
        \multirow{4}{*}{Bench} & MeshGPT~\cite{siddiqui2024meshgpt} & 56.06 & 8.44 & 58.33 & 28.34  & 66.30  & 9.45  & 
        \multirow{4}{*}{Lamp} & MeshGPT~\cite{siddiqui2024meshgpt} & 43.90 & 20.82 & 60.37 & \underline{36.21}  & 73.21  & 6.04     \\
        & PivotMesh~\cite{weng2024pivotmesh} & \textbf{59.09} & 8.25 & \underline{48.48} & \textbf{25.76}  & 64.48  & 5.17  & 
        & PivotMesh~\cite{weng2024pivotmesh} & \underline{50.00} & \underline{19.17} & \underline{56.71} & 39.75  & 67.76  & 7.09   \\
        & MeshXL*~\cite{chen2024meshxl} & \textbf{59.09} & \textbf{7.74} & 53.79 & \underline{26.37}  & \textbf{19.30}  & \underline{3.44}  & 
        & MeshXL*~\cite{chen2024meshxl} & 42.68 & 21.64 & 63.41 & \textbf{35.96}  & \textbf{62.46}  & \underline{5.32} \\
        & \textbf{Ours} & \underline{57.58} & \underline{7.90} & \textbf{50.76} & 27.17  & \underline{59.83}  & \textbf{1.53}  & 
        & \textbf{Ours} & \textbf{62.20} & \textbf{18.69} & \textbf{48.17} & 37.33  & \underline{62.81}  & \textbf{2.17} \\
        \midrule
      \end{tabular}
    }
    \vspace{-0.2cm}
    \caption{\textbf{Quantitative results on ShapeNet dataset.} MeshCraft outperforms the baselines on shape quality, visual and compactness metrics. MMD values are multiplied by $10^3$. COV and 1-NNA are scaled by $10^2$. * stands for using the released pre-trained models.}
    \label{tab:shapenet_quantitative}
    \vspace{-1.0cm}
  \end{center}
\end{table*}

\subsection{Mesh generation with the flow-based DiT}
\label{sec:dit}
As illustrated in the right part of \cref{fig:pipeline}, the face-level tokens are fed into the flexible diffusion transformer for training. However, standard SiT~\cite{ma2024sit} does not support direct training with token sequences of variable lengths. To address this, we introduce several techniques to adapt the architecture for mesh generation.

First, it is essential that the input tokens are of the same length. By padding the token sequences to match the length of the longest mesh sequence in the batch, the model can be trained using mesh sequences of varying lengths. The corresponding masks for these samples are also provided to the model, functioning as attention masks within the SiT blocks and guiding the unpadding process. RoPE~\cite{su2024roformer} is applied to keys/values in attention layer. And the attention scores are computed as follows:
\begin{equation}
\label{eq:attn}
    \mathrm{Softmax}\left( \frac{\mathbf{Q} \mathbf{K}^T}{\sqrt{d}} + M \right)
\end{equation}
where \(\mathbf{Q}\) and \(\mathbf{K}\) represent the queries and keys, respectively, and \(d\) is their dimension. The value of $M$ is set to be 0 for noised tokens and \(-\infty\) for padding tokens. Masks are also applied to exclude padding tokens before calculating losses, ensuring that generated tokens align with the input tokens. (refer to the supplementary for our detailed architectures)

In addition to masking techniques, we explicitly use the number of faces \(c_{f}\) as guidance. Following the class conditioning approach implemented with adaLN-Zero blocks in DiT~\cite{peebles2023scalable}, we embed the number of faces using an embedding layer and add it to the timestep embeddings. This modification provides additional information and models a more accurate distribution of meshes according to the targeted number of faces, significantly enhancing generation quality when combined with classifier-free guidance (CFG)~\cite{ho2022classifier}:
\begin{equation}
    \widetilde{v}_t=v_\theta(z_t, \emptyset)+w\cdot (v_\theta(z_t, c_{f}) - v_\theta(z_t, \emptyset))
\end{equation}
where \(\theta\) and \(z_t\) denote the network parameters and noised tokens, respectively. \(\widetilde{v}_t\) represents the estimated velocity at each de-noising step. For conditioning the mesh geometry (e.g., from images or texts), we employ cross-attention modules to inject the corresponding features. Thus, for image-conditioned generation (similar to text conditions), we have two conditions: the input image \(c_i\) and the assigned number of faces \(c_{f}\). We find that the generation process benefits from multiple different CFG weights. Therefore, our modified estimation equation for the multiple conditions \(c_f\) and \(c_i\) can be expressed as:

\begin{equation}
\begin{split}
    \widetilde{v}_t&=v_\theta(z_t, \emptyset, \emptyset)\\
            &= +w_1\cdot (v_\theta(z_t, c_{f}, \emptyset) - v_\theta(z_t, \emptyset, \emptyset)) \\
            &= +w_2\cdot (v_\theta(z_t, c_{f}, c_i) - v_\theta(z_t, c_f, \emptyset))
\end{split}
\end{equation}
In \cref{sec:ablation_cfg}, we present the effects of CFG weights on the generated meshes.

Additionally, to ensure more stable training, we introduce sandwich normalization~\cite{ding2021cogview}, replace the MLP with SwiGLU, and employ QK-norm~\cite{dehghani2023scaling} techniques inspired by previous works~\cite{touvron2023llama, zhuo2024lumina}. The QK-norm is crucial for stabilizing the training of transformer models, especially when the token length gets more flexible and longer.
Benefiting from the elaborate architecture and excellent properties of rectified flow's linear sampling, training becomes more efficient and stable. We also draw inspiration from observations during the noising process to prioritize intermediate steps of diffusion, as the middle and final stages are more challenging during training. To alleviate this, we adopt the logit-normal sampling strategy from SD3~\cite{esser2024scaling} to adjust the sampling weights:
\begin{equation}
\label{eq:sample}
    \pi_{\mathrm{ln}}(t;m,s)=\frac{1}{s\sqrt{2\pi}t(1-t)}\exp \left( -\frac{(\log (t/(1-t))-m)^2}{2s^2} \right)
\end{equation}
where \(m\) and \(s\) represent the location and scale parameters, respectively.

Finally, we utilize an MSE loss \(\mathcal{L}_{Diff}\) to predict the velocity \(v\) using our model.

\begin{figure}[!t]
\begin{center}
\centerline{\includegraphics[width=\linewidth]{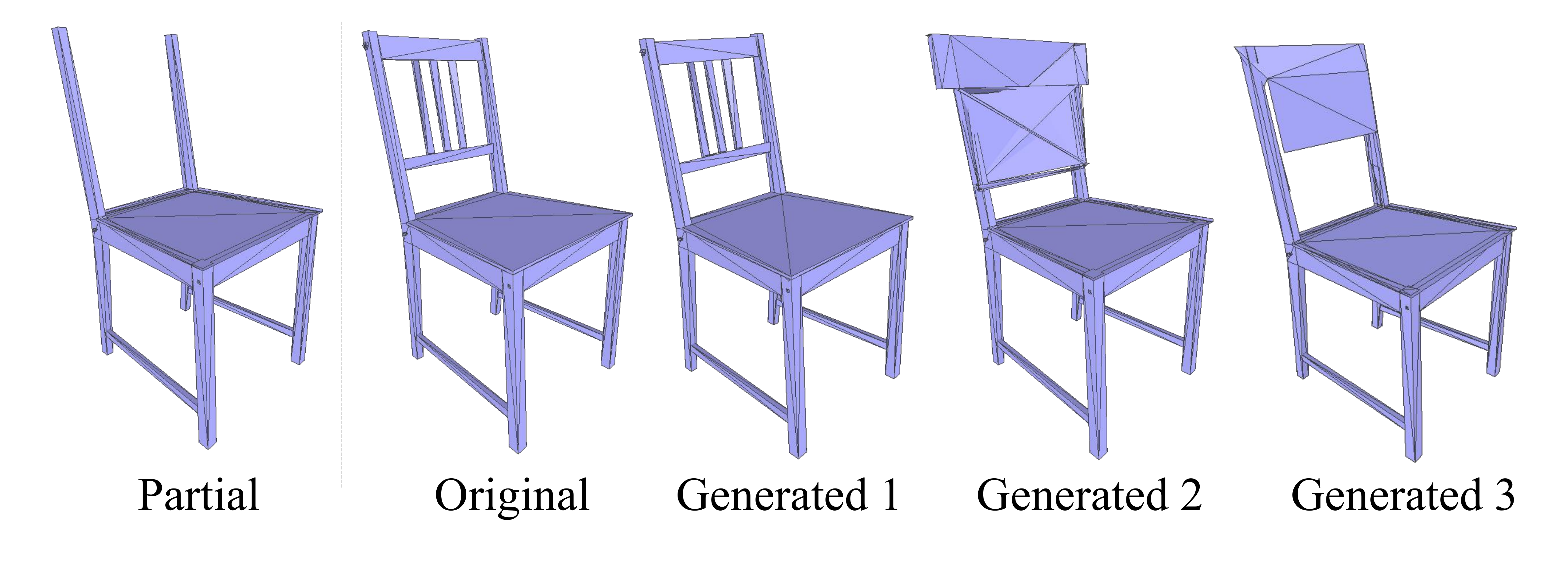}}
\vspace{-0.4cm}
\caption{\textbf{Mesh completion results.} Given some partial observation of a mesh, MeshCraft can produce diverse completed results.}
\label{fig:completion}
\end{center}
\vspace{-1.3cm}
\end{figure}

%% file: sec/5_experiments.tex
\section{Experiments}
\label{sec:exp}
\begin{figure*}[!t]
\begin{center}
\centerline{\includegraphics[width=\linewidth]{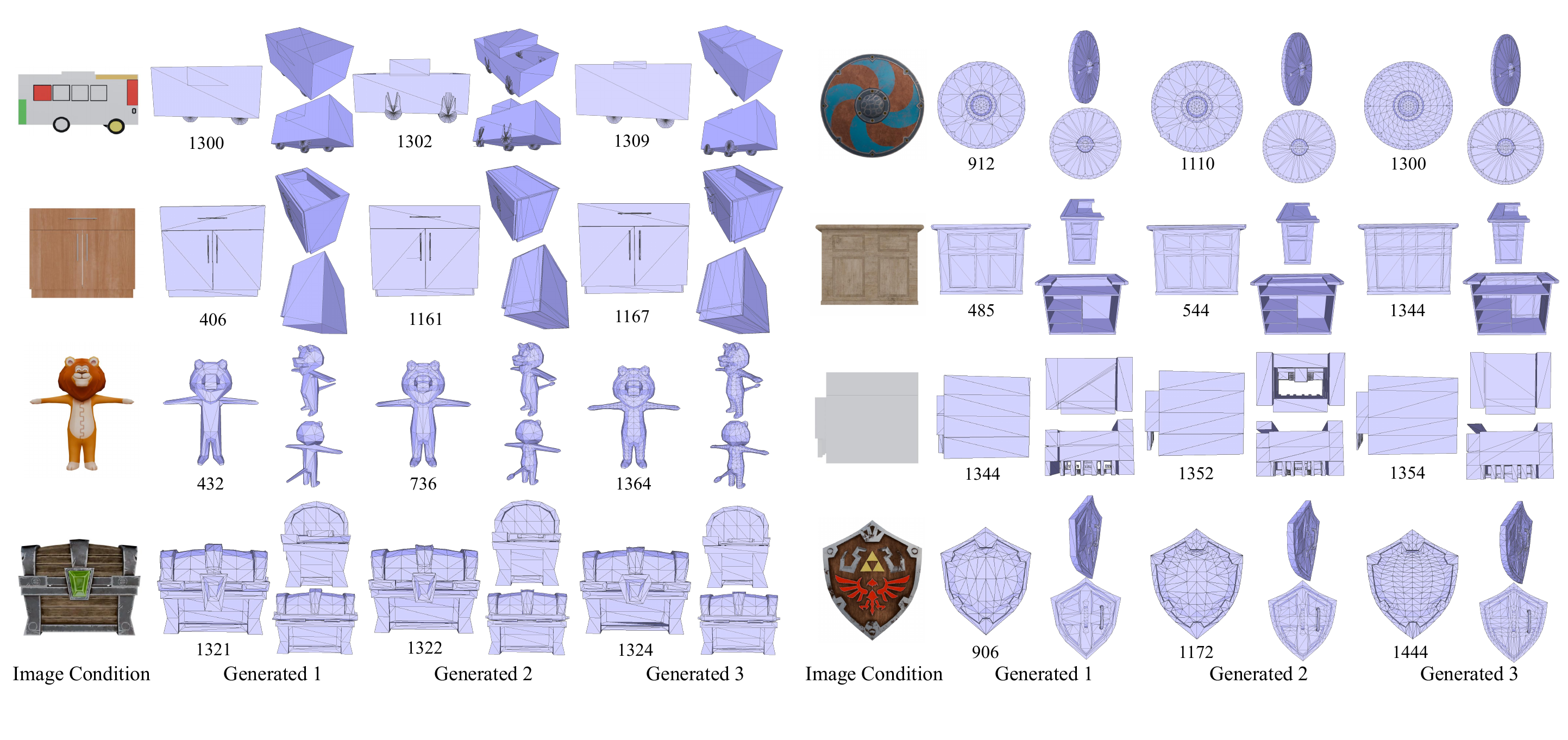}}
\vspace{-0.8 cm}
\caption{\textbf{Generation diversity on Objaverse dataset.} The number below each asset represents for the face number of it. MeshCraft is about to produce diverse samples with different seeds and face numbers.}
\label{fig:objaverse_diversity}
\end{center}
\vspace{-1.2cm}
\end{figure*}

We trained our model on two datasets: ShapeNet~\cite{chang2015shapenet} and Objaverse~\cite{deitke2023objaverse}. The model on ShapeNet demonstrates the effectiveness of our method (see \cref{fig:shapenet_qualitative} and \cref{tab:shapenet_quantitative} for the main results), which achieves state-of-the-art performance compared with prior works. Furthermore, experiments (\cref{fig:objaverse_diversity}) conducted on Objaverse show that MeshCraft is able to generate diverse samples of different face numbers conditioned on the same single image, which shows its potential to be a practical mesh generator in larger-scale scenarios.

\subsection{Experiment Settings}

\subsubsection{Datasets}
Following the previous convention~\cite{siddiqui2024meshgpt, weng2024pivotmesh, chen2024meshxl} for preprocessing, we apply planar decimation to meshes with more than $800$ faces and use further filtering by comparing the Hausdorff distance~\cite{huttenlocher1993comparing} between decimated and the original meshes with a pre-set threshold $\sigma_{hausdorff}$ for the ShapeNet dataset. As for the Objaverse dataset, we select assets whose number of faces is in $[1024, 1536]$ to train our image-conditioned model. The dataset size is about 10k and 65k, respectively. We split the training set and validation set in 10:1 and 100:1. The coordinate space resolution is set as 128 and 256 for ShapeNet and Objaverse dataset, respectively. We normalize the coordinate range into $[-1, 1]$ and augment the data using random scaling on each axis (from 0.95 to 1.05) and random rotations for the reconstruction stage to improve auto-encoder's robustness. Across all experiments, we use triangular meshes (each face $f_i$ consists of 3 vertices) as training data for fair comparisons with baselines.

\subsubsection{Evaluation}
We choose three recent baselines to compare the results on ShapeNet dataset, including MeshGPT~\cite{siddiqui2024meshgpt}, PivotMesh~\cite{weng2024pivotmesh} and MeshXL~\cite{chen2024meshxl}. For evaluating the reconstruction quality, we follow previous works~\cite{weng2024pivotmesh, siddiqui2024meshgpt, alliegro2023polydiff} to use two metrics, triangle accuracy and l2 distance. 
We evaluate the methods following previous mesh generation works~\cite{weng2024pivotmesh}. The metrics include Minimum Matching Distance (MMD), Coverage (COV), 1-Nearest-Neighbor Accuracy (1-NNA), Jensen-Shannon Divergence (JSD) for measuring 3D geometry, FID and KID for visual perceptions. For COV, higher is better; for 1-NNA, 50\% is the optimal; for the rest of metrics, lower is better. Following PivotMesh~\cite{weng2024pivotmesh}, we use a Chamfer Distance measure for computing these metrics on 1024-dim point clouds uniformly sampled from meshes.
For each method, we generate 1000 samples to calculate the quantitative results.

\subsubsection{Implementation Details}
For transformer-styled blocks in the auto-encoder, we set the encoder part as 12 layers with a hidden size of 768, and 18 layers with a hidden size of 384 for the decoder part. The channel dimension for face-level tokens is set to 8 for the balance of reconstruction quality and compression capability. For the diffusion model, we adopt a 24-layer transformer with a hidden size of 864, which has similar number of parameters compared with baselines. For the implementation of baselines, we adopt the public code\footnote{\url{https://github.com/lucidrains/meshgpt-pytorch}} to reimplement MeshGPT~\cite{siddiqui2024meshgpt}, and official training code of PivotMesh~\cite{weng2024pivotmesh}. For MeshXL~\cite{chen2024meshxl}, we use their released 350M pre-trained models for evaluation. All models are trained following the settings claimed in the original paper. Without further specification, we generate meshes in the distribution of data's face numbers for MeshCraft and follow the default settings of baselines. For auto-encoders, we trained about 2 days on an 8$\times$A100 80GB machine with a batch size of 8. For diffusion transformers, we train for 3 days on the ShapeNet dataset and for around 3 weeks on the Objaverse dataset. During the training of diffusion transformers, we use bf16 mixed precision to accelerate the training process. All the generated results are sampled with the 50-step Euler method.

\subsection{Experiment Results}
\subsubsection{Results on ShapeNet dataset}

Firstly, we evaluate our method on the most commonly used benchmark, ShapeNet, focusing on four different categories: chair, table, bench, and lamp. \cref{tab:recon_shapenet} shows that our continuous latent space VAE achieves competitive reconstruction performance compared with baselines that adopt an advanced vector quantization technique, RVQ~\cite{zeghidour2021soundstream}, to discretize vertex coordinates into indices of a codebook. Furthermore, the number of tokens in face sequences decreases ninefold compared with prior works, which not only reduces memory requirements but also speeds up the subsequent generation process.
For the comparisons in the generation part, we follow the previous setting~\cite{weng2024pivotmesh, siddiqui2024meshgpt, alliegro2023polydiff, chen2024meshxl}, first pretraining our model on a mixed dataset composed of the four categories and subsequently fine-tuning each of them separately to perform comparisons from both qualitative and quantitative aspects. For the CFG weight $w$ of the face number condition, we set $w = 8.0$ for better generation quality. As shown in \cref{fig:shapenet_qualitative}, our method can produce diverse and high-quality 3D meshes. As \cref{tab:shapenet_quantitative} demonstrates, our method significantly outperforms the baselines, performing better on all three metrics. In \cref{fig:completion}, we also demonstrate the completion ability of our model by adapting the operation from an image inpainting work~\cite{lugmayr2022repaint}. Compared with autoregressive methods, our diffusion-based method achieves state-of-the-art generation results while maintaining a superior generation speed. Specifically, as shown in \cref{fig:teaser}(b) and (c), our method decreases the token number by up to 9 times and speeds up generation by 35 times, demonstrating the advantages of diffusion-based models.

\subsubsection{Results on Objaverse dataset}
An image-conditioned model is also trained to demonstrate the effectiveness of face number control on large-scale dataset Objaverse. To enhance the capability of capturing details in the image condition, we use DINOv2 ViT-L/14~\cite{oquab2023dinov2} as the image feature extractor and fine-tuned with 3D assets whose front view occupies more than 20\% of the frame. For the CFG weights $w1,w2$, which respectively controls the face number and the input image condition, we set them as $w1=1.0, w2=5.0$.
\cref{fig:objaverse_diversity} displays that our model has the potential for the capability of diverse generation on the large-scale dataset, which can be also combined well with existing mature conditional techniques~\cite{ho2022classifier}. Notably, increasing the face number primarily enhances fine local details rather than uniformly refining the entire shape. For example, in the third row, the tiger's facial features become more detailed, whereas broader regions such as the torso show only slight improvements.

\begin{table}[t]
\begin{center}
\resizebox{\linewidth}{!}{
\begin{tabular}{ccc}
\toprule
\bf Method & \bf Tri. Accu.$(\%)\uparrow$ & \bf L2 Dist.$(\times 10^{-2})\downarrow$ \\
\hline
KL (4-dim) & 90.41 & 0.63 \\
 \hline 
KL (8-dim) & 99.66 & 0.12  \\
\hline
RVQ~\cite{zeghidour2021soundstream}    & 65.12 & 8.63 \\
\bottomrule
\end{tabular}}
\end{center}
\vspace{-0.5cm}
\caption{\textbf{Reconstruction performance study on Objverse.}}
\label{tab:recon}
\vspace{-0.7cm}
\end{table}

\subsection{Ablation Studies}
\subsubsection{Comparisons of auto-encoder selections}
\label{sec:ablation_ae}
At the reconstruction stage, we compare auto-encoders operating in different continuous and discrete latent spaces. As shown in \cref{tab:recon}, KL regularization with 8-dimensional token channels yields better performance, striking a good balance between reconstruction quality and compression capability compared to using 4-dimensional channels. We believe that using an excessively high compression ratio leads to severe information loss. Additionally, we compare the continuous tokenizer with the discrete one by replacing the compression component using KL regularization with residual vector quantization (RVQ)~\cite{zeghidour2021soundstream}. It is worth mentioning that discrete tokenizers, limited by their codebook size, can cause significant information loss when compressing variable data. In contrast, models with continuous tokens produce meshes of higher quality.

\subsubsection{Effects of CFG weights}
\label{sec:ablation_cfg}
\begin{figure}[!t]
\begin{center}
\centerline{\includegraphics[width=1.05\linewidth]{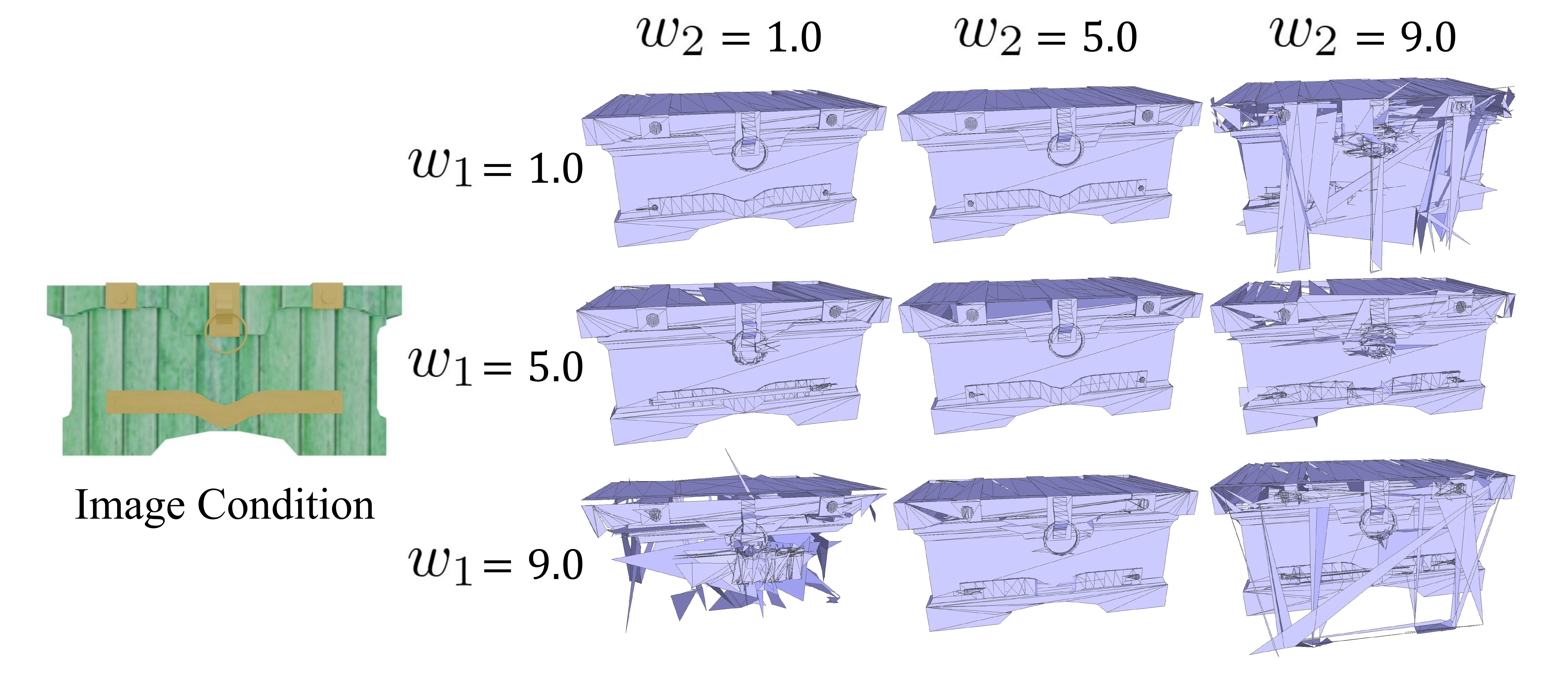}}
\vspace{-0.4cm}
\caption{\textbf{CFG weights over face number and image conditions.}  $w_1$ controls on the condition of the face number $c_f$, while $w_2$ yields weights further over the single-image condition $c_i$.}
\label{fig:cfg_effect}
\end{center}
\vspace{-1.2cm}
\end{figure}

We also investigate the effects of CFG weights on $w$ and $w_1,w_2$, respectively. For the weight $w$ controlling the face number, \cref{tab:cfg_shapenet} shows that as $w$ increases, the quantitative metrics improve, reaching a peak when $w$ is around 8.0. 
For multiple conditions (single images $c_i$ and the number of faces $c_f$), \cref{fig:cfg_effect} provides a visualization of the effects corresponding to different scales of $w_1, w_2$. The best result appears when $w1=1.0, w2=5.0$. These two sets of experiments show that low weights can result in weak conditional control, while excessive weights are detrimental to the results.

%


\begin{table}[t]
\begin{center}
    
    \begin{tabular}{cccc}
    \toprule
\bf CFG & \bf COV$\uparrow$  & \bf MMD$\downarrow$ & \bf 1-NNA  \\
\hline
0.0 & 30.70 & 12.61 & 84.75 \\
 \hline 
1.0 & 38.20 & 11.34 & 75.00 \\
\hline
2.0 & 44.80 & 10.72 & 67.65 \\
 \hline 
3.0 & 48.60 & 10.33 & 61.05 \\
\hline
4.0 & 50.80 & 9.89 & 58.40 \\
 \hline 
5.0 & 51.70 & 10.05 & 57.55 \\
\hline
6.0 & 51.00 & 9.97 & \textbf{56.55}  \\
 \hline 
7.0 & 51.70 & \textbf{9.65} & \underline{56.70} \\
\hline
8.0 & \textbf{53.30} & 9.90 & 56.85 \\
 \hline 
9.0 & \underline{52.50} & \underline{9.85} & 56.75 \\
\hline
10.0 & 52.30 & 9.86 & 57.90 \\
\bottomrule
\end{tabular}
\vspace{-0.2cm}
\caption{\textbf{Effects of CFG weights over the face number condition to mesh generation results on ShapeNet dataset.}}
\label{tab:cfg_shapenet}
\end{center}
\vspace{-0.8cm}
\end{table}

\subsection{Limitations}
Though MeshCraft shows promising results, there are still some limitations: (1) The extrapolation capability of our diffusion model is limited due to the face number embedder we use is learnable, and objects with unseen face numbers cannot be produced directly; (2) When the domain of images and assigned face numbers is far from the training domain, MeshCraft fails to generate completed meshes. We will further explore more generalizable models with improved training strategies and architectures.

%% file: sec/6_conclusions.tex
\section{Conclusion}
\label{sec:conclusion}
In this paper, we introduce a novel method, namely MeshCraft, for generating ready-to-use 3D meshes with high efficiency and controllability. Regarding meshes as face-level sequences, we first compress them into continuous tokens and subsequently generate the tokens with a flow-based diffusion transformer. Our method demonstrates superior speed (35 $\times$ speed up) and shows competitive performance in both qualitative and quantitative experiments. MeshCraft shows the potential to alleviate artists from time-consuming manual work.

%% file: sec/X_suppl.tex
\clearpage
\setcounter{page}{1}
\maketitlesupplementary

\begin{figure}[!t]
\begin{center}
\centerline{\includegraphics[width=\linewidth]{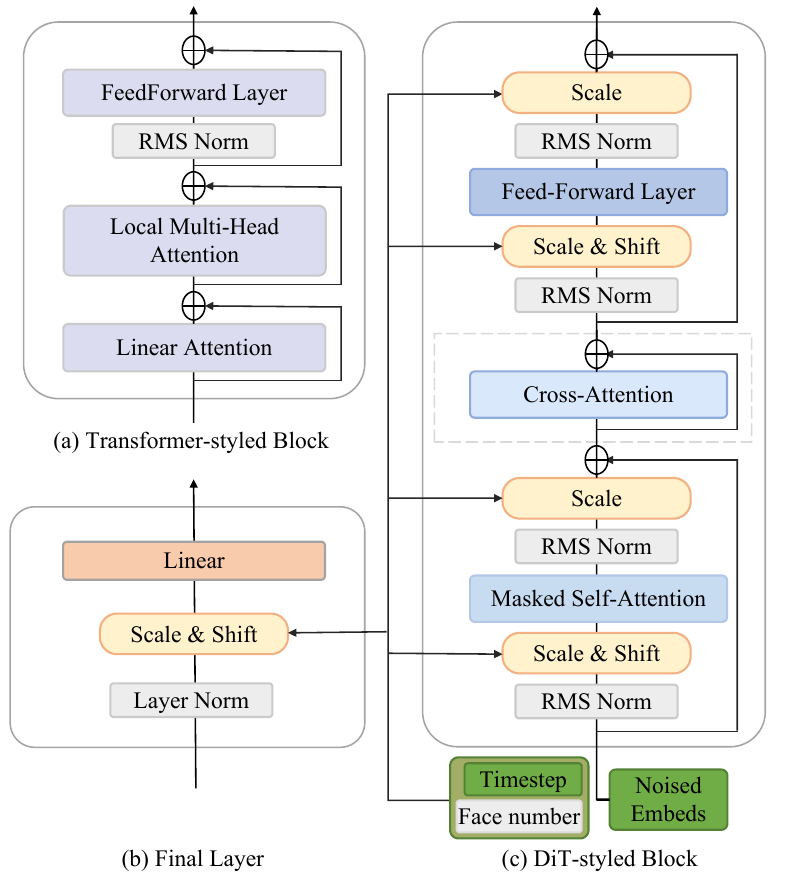}}
\caption{\textbf{Architectures} of the transformer-styled block and well-designed blocks in the DiT.}
\label{fig:arch}
\end{center}
\end{figure}

\section{Additional details of the model design}
\subsection{Detailed architectures of modules}
The proposed model architecture comprises intricate transformer-styled blocks with carefully designed components. As illustrated in~\cref{fig:arch}(a), the transformer-styled block is composed of two attention layers, followed by an RMS normalization and a feed-forward layer. \cref{fig:arch}(b) and (c) provide a detailed visualization of our DiT-styled block and the final layer in the diffusion transformer, highlighting the nuanced design choices that contribute to the model's performance.

\subsection{Logit-normal sampling}
\begin{figure}[!t]
\begin{center}
\centerline{\includegraphics[width=1\linewidth]{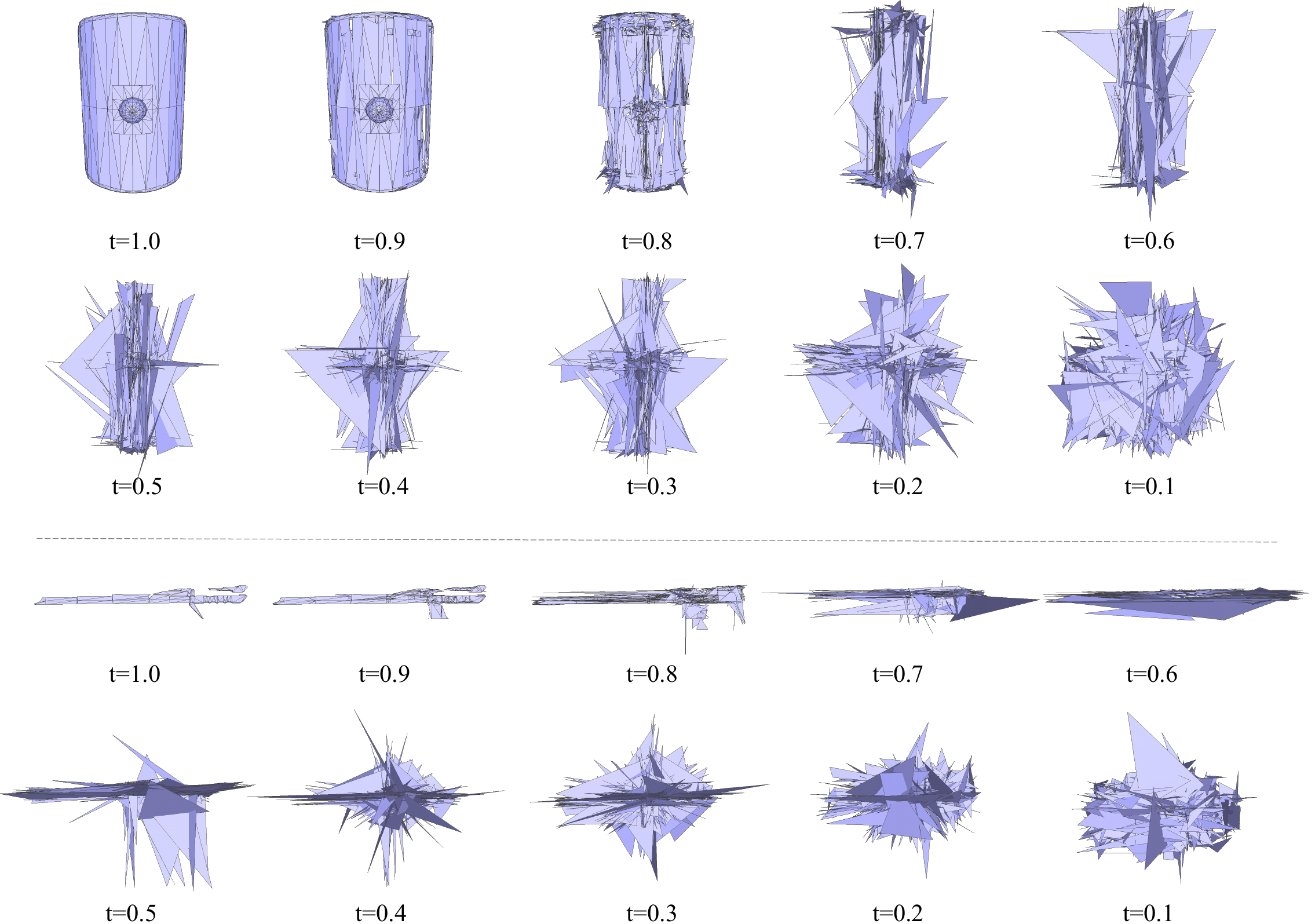}}
\caption{\textbf{Process of adding noises.} The complete mesh is gradually transformed into noises from standard normal distribution from $t=1$ to $t=0$.}
\label{fig:noise}
\end{center}
\end{figure}

Drawing inspiration from the training methodology of SD3~\cite{esser2024scaling}, we inspect the diffusion process of noises to the latent tokens, visualizing in \cref{fig:noise}. The visualization in \cref{fig:noise} reveals critical insights into the mesh generation process, demonstrating that intermediate and final diffusion steps play a pivotal role in generating complete meshes. Consequently, we implemented a logit-normal sampling approach to emphasize these crucial stages of the generation process. By setting the distribution parameters to $m=0.5$ and $s=1.0$, informed by empirical results from SD3, we effectively prioritize sampling in the most informative regions of the diffusion process.

\begin{figure}[!t]
\begin{center}
\centerline{\includegraphics[width=1.1\linewidth]{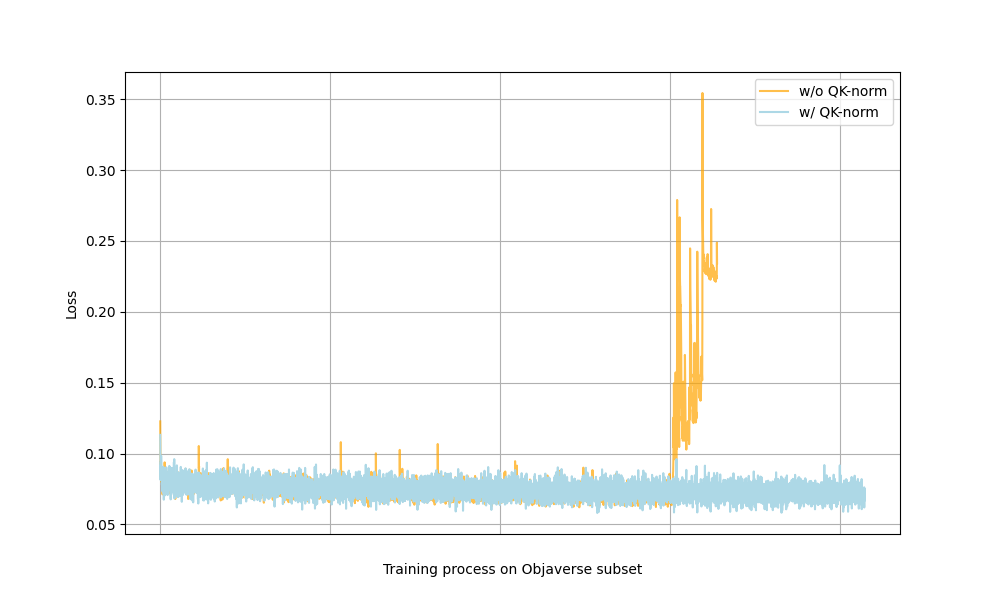}}
\caption{\textbf{Loss during training on Objaverse dataset.} The QK-norm is of vital importance for stabilizing the training process.}
\label{fig:objaverse_loss}
\end{center}
\end{figure}

\subsection{The importance of QK-norm}
Recent advancements in transformer research~\cite{dehghani2023scaling, lu2024fit, zhuo2024lumina} have highlighted the inherent challenges of training large-parameter models with flexible data sequences. Our mesh generation experiments on the Objaverse dataset corroborated these observations, revealing significant training instabilities. To address this critical issue, we implemented the QK-norm technique, a proven strategy for mitigating training volatility. Formally, we modified the attention scores as follows:
\begin{equation}
    \mathrm{Softmax}\left( \frac{\mathrm{LN}(\mathbf{Q}) \mathrm{LN}(\mathbf{K})^T}{\sqrt{d}} + M \right)
\end{equation}
By applying LayerNorm to the query and key matrices, we effectively stabilize the attention mechanism. \cref{fig:objaverse_loss} demonstrates that QK-norm helps to stabilize the process.

\begin{figure*}[!t]
\begin{center}
\centerline{\includegraphics[width=\linewidth]{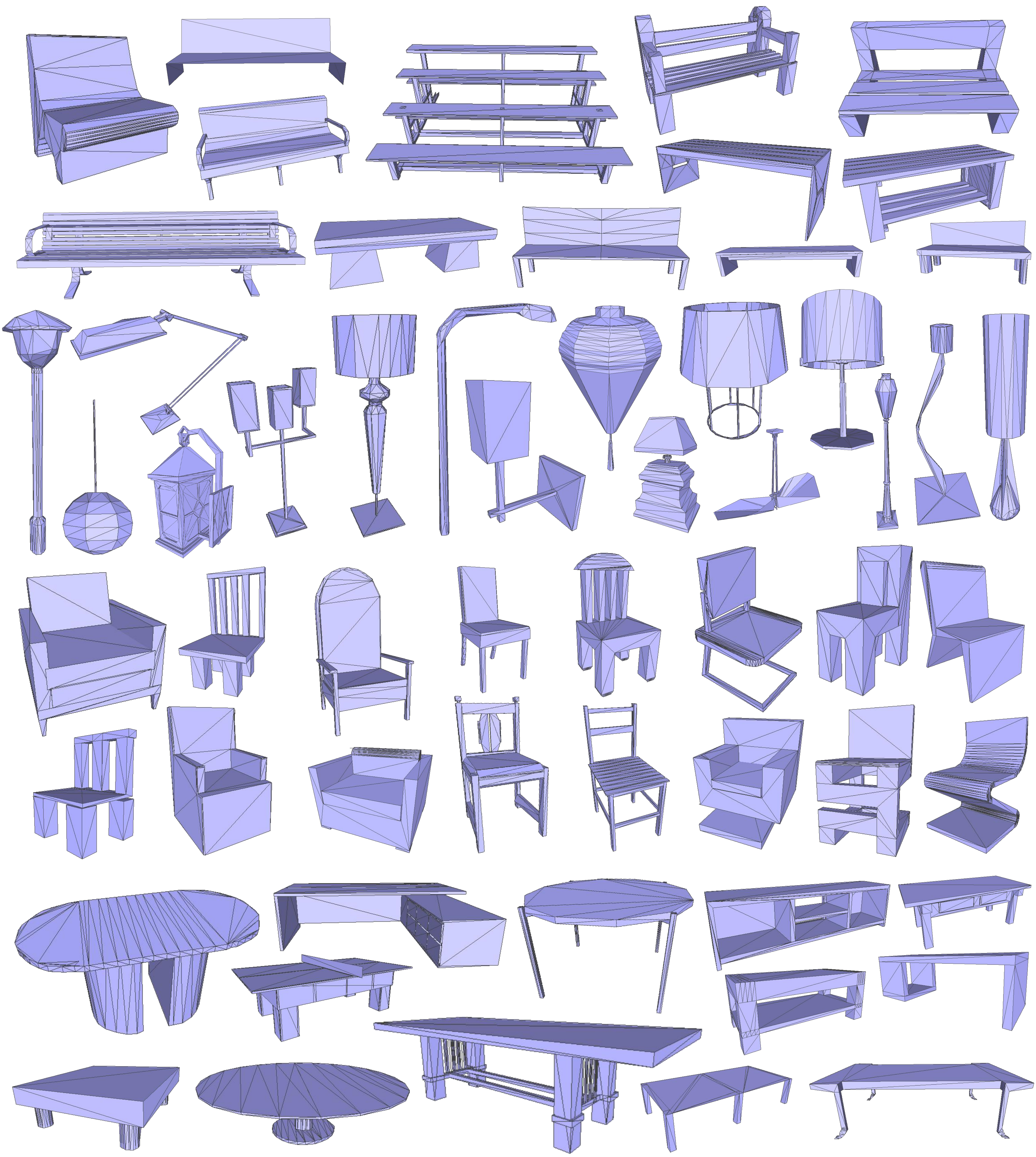}}
\caption{\textbf{Generation gallery on ShapeNet.} Additional results on the subset of bench, lamp, chair and table.}
\label{fig:shapenet_gallery}
\end{center}
\end{figure*}

\begin{figure*}[!t]
\begin{center}
\centerline{\includegraphics[width=\linewidth]{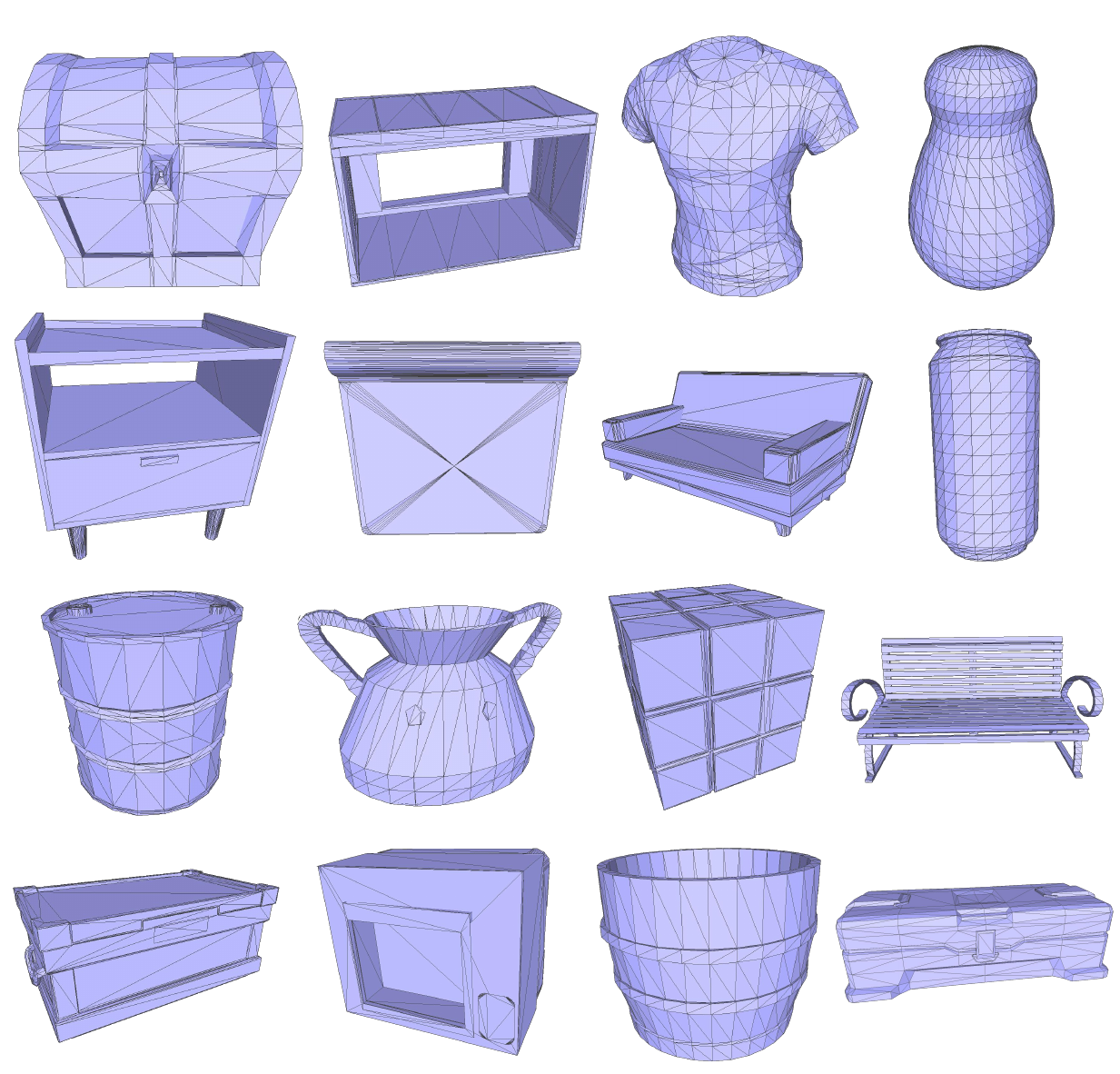}}
\caption{\textbf{Additional generation results on Objaverse.}}
\label{fig:objaverse_gallery}
\end{center}
\end{figure*}

\begin{figure*}[!t]
\begin{center}
\centerline{\includegraphics[width=\linewidth]{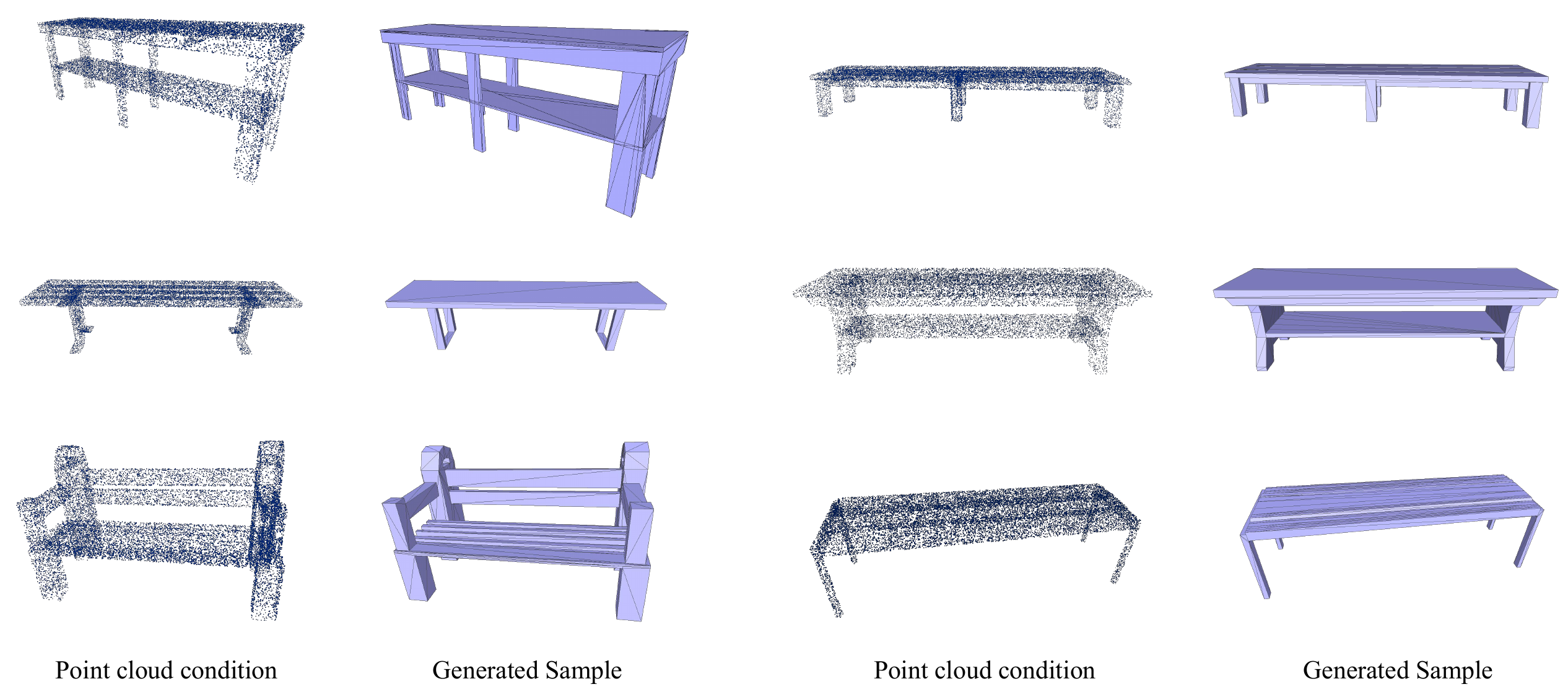}}
\caption{\textbf{Point cloud conditioned generation on the ShapeNet bench dataset.}}
\label{fig:pc_cond_gallery}
\end{center}
\end{figure*}

\section{Additional quantitative results}
We present comprehensive visualization of generation results across multiple datasets, showcasing the versatility of our approach in \cref{fig:shapenet_gallery}, \cref{fig:objaverse_gallery}, and \cref{fig:pc_cond_gallery}. To demonstrate the model's adaptability, we also extended our method to point cloud conditioning on the ShapeNet bench dataset. Our implementation leverages a pre-trained point cloud encoder inspired by the Michelangelo~\cite{zhao2024michelangelo} architecture. We integrated the point cloud features into our model using cross-attention techniques, analogous to image feature injection. Specifically, we employed linear projections to seamlessly adapt and align the point cloud representations with our model's internal feature space. Refer to \cref{fig:pc_cond_gallery} for the results.

%% file: main.bbl
\begin{thebibliography}{42}
\providecommand{\natexlab}[1]{#1}
\providecommand{\url}[1]{\texttt{#1}}
\expandafter\ifx\csname urlstyle\endcsname\relax
  \providecommand{\doi}[1]{doi: #1}\else
  \providecommand{\doi}{doi: \begingroup \urlstyle{rm}\Url}\fi

\bibitem[Alliegro et~al.(2023)Alliegro, Siddiqui, Tommasi, and Nie{\ss}ner]{alliegro2023polydiff}
Antonio Alliegro, Yawar Siddiqui, Tatiana Tommasi, and Matthias Nie{\ss}ner.
\newblock Polydiff: Generating 3d polygonal meshes with diffusion models.
\newblock \emph{arXiv preprint arXiv:2312.11417}, 2023.

\bibitem[Chang et~al.(2015)Chang, Funkhouser, Guibas, Hanrahan, Huang, Li, Savarese, Savva, Song, Su, et~al.]{chang2015shapenet}
Angel~X Chang, Thomas Funkhouser, Leonidas Guibas, Pat Hanrahan, Qixing Huang, Zimo Li, Silvio Savarese, Manolis Savva, Shuran Song, Hao Su, et~al.
\newblock Shapenet: An information-rich 3d model repository.
\newblock \emph{arXiv preprint arXiv:1512.03012}, 2015.

\bibitem[Chen et~al.(2024{\natexlab{a}})Chen, Wu, Luo, Xie, Paul, Luo, Zhao, and Li]{chen2024pixart}
Junsong Chen, Yue Wu, Simian Luo, Enze Xie, Sayak Paul, Ping Luo, Hang Zhao, and Zhenguo Li.
\newblock Pixart-$\{$$\backslash$delta$\}$: Fast and controllable image generation with latent consistency models.
\newblock \emph{arXiv preprint arXiv:2401.05252}, 2024{\natexlab{a}}.

\bibitem[Chen et~al.(2024{\natexlab{b}})Chen, Chen, Pang, Zeng, Cheng, Fu, Yin, Wang, Wang, Zhang, et~al.]{chen2024meshxl}
Sijin Chen, Xin Chen, Anqi Pang, Xianfang Zeng, Wei Cheng, Yijun Fu, Fukun Yin, Yanru Wang, Zhibin Wang, Chi Zhang, et~al.
\newblock Meshxl: Neural coordinate field for generative 3d foundation models.
\newblock \emph{arXiv preprint arXiv:2405.20853}, 2024{\natexlab{b}}.

\bibitem[Chen et~al.(2024{\natexlab{c}})Chen, He, Huang, Ye, Chen, Tang, Chen, Cai, Yang, Yu, et~al.]{chen2024meshanything}
Yiwen Chen, Tong He, Di Huang, Weicai Ye, Sijin Chen, Jiaxiang Tang, Xin Chen, Zhongang Cai, Lei Yang, Gang Yu, et~al.
\newblock Meshanything: Artist-created mesh generation with autoregressive transformers.
\newblock \emph{arXiv preprint arXiv:2406.10163}, 2024{\natexlab{c}}.

\bibitem[Dehghani et~al.(2023)Dehghani, Djolonga, Mustafa, Padlewski, Heek, Gilmer, Steiner, Caron, Geirhos, Alabdulmohsin, et~al.]{dehghani2023scaling}
Mostafa Dehghani, Josip Djolonga, Basil Mustafa, Piotr Padlewski, Jonathan Heek, Justin Gilmer, Andreas~Peter Steiner, Mathilde Caron, Robert Geirhos, Ibrahim Alabdulmohsin, et~al.
\newblock Scaling vision transformers to 22 billion parameters.
\newblock In \emph{International Conference on Machine Learning}, pages 7480--7512. PMLR, 2023.

\bibitem[Deitke et~al.(2023)Deitke, Schwenk, Salvador, Weihs, Michel, VanderBilt, Schmidt, Ehsani, Kembhavi, and Farhadi]{deitke2023objaverse}
Matt Deitke, Dustin Schwenk, Jordi Salvador, Luca Weihs, Oscar Michel, Eli VanderBilt, Ludwig Schmidt, Kiana Ehsani, Aniruddha Kembhavi, and Ali Farhadi.
\newblock Objaverse: A universe of annotated 3d objects.
\newblock In \emph{Proceedings of the IEEE/CVF Conference on Computer Vision and Pattern Recognition}, pages 13142--13153, 2023.

\bibitem[Ding et~al.(2021)Ding, Yang, Hong, Zheng, Zhou, Yin, Lin, Zou, Shao, Yang, et~al.]{ding2021cogview}
Ming Ding, Zhuoyi Yang, Wenyi Hong, Wendi Zheng, Chang Zhou, Da Yin, Junyang Lin, Xu Zou, Zhou Shao, Hongxia Yang, et~al.
\newblock Cogview: Mastering text-to-image generation via transformers.
\newblock \emph{Advances in neural information processing systems}, 34:\penalty0 19822--19835, 2021.

\bibitem[Esser et~al.(2024)Esser, Kulal, Blattmann, Entezari, M{\"u}ller, Saini, Levi, Lorenz, Sauer, Boesel, et~al.]{esser2024scaling}
Patrick Esser, Sumith Kulal, Andreas Blattmann, Rahim Entezari, Jonas M{\"u}ller, Harry Saini, Yam Levi, Dominik Lorenz, Axel Sauer, Frederic Boesel, et~al.
\newblock Scaling rectified flow transformers for high-resolution image synthesis.
\newblock In \emph{Forty-first International Conference on Machine Learning}, 2024.

\bibitem[Fan et~al.(2024)Fan, Li, Qin, Li, Sun, Rubinstein, Sun, He, and Tian]{fan2024fluid}
Lijie Fan, Tianhong Li, Siyang Qin, Yuanzhen Li, Chen Sun, Michael Rubinstein, Deqing Sun, Kaiming He, and Yonglong Tian.
\newblock Fluid: Scaling autoregressive text-to-image generative models with continuous tokens.
\newblock \emph{arXiv preprint arXiv:2410.13863}, 2024.

\bibitem[Hao et~al.(2024)Hao, Romero, Lin, and Liu]{hao2024meshtron}
Zekun Hao, David~W Romero, Tsung-Yi Lin, and Ming-Yu Liu.
\newblock Meshtron: High-fidelity, artist-like 3d mesh generation at scale.
\newblock \emph{arXiv preprint arXiv:2412.09548}, 2024.

\bibitem[He et~al.(2025)He, Chen, Peng, Huang, Li, Huang, Yuan, Ouyang, and He]{he2025gvgen}
Xianglong He, Junyi Chen, Sida Peng, Di Huang, Yangguang Li, Xiaoshui Huang, Chun Yuan, Wanli Ouyang, and Tong He.
\newblock Gvgen: Text-to-3d generation with volumetric representation.
\newblock In \emph{European Conference on Computer Vision}, pages 463--479. Springer, 2025.

\bibitem[Ho and Salimans(2022)]{ho2022classifier}
Jonathan Ho and Tim Salimans.
\newblock Classifier-free diffusion guidance.
\newblock \emph{arXiv preprint arXiv:2207.12598}, 2022.

\bibitem[Ho et~al.(2020)Ho, Jain, and Abbeel]{ho2020denoising}
Jonathan Ho, Ajay Jain, and Pieter Abbeel.
\newblock Denoising diffusion probabilistic models.
\newblock \emph{Advances in neural information processing systems}, 33:\penalty0 6840--6851, 2020.

\bibitem[Huttenlocher et~al.(1993)Huttenlocher, Klanderman, and Rucklidge]{huttenlocher1993comparing}
Daniel~P Huttenlocher, Gregory~A. Klanderman, and William~J Rucklidge.
\newblock Comparing images using the hausdorff distance.
\newblock \emph{IEEE Transactions on pattern analysis and machine intelligence}, 15\penalty0 (9):\penalty0 850--863, 1993.

\bibitem[Kerbl et~al.(2023)Kerbl, Kopanas, Leimk{\"u}hler, and Drettakis]{kerbl20233d}
Bernhard Kerbl, Georgios Kopanas, Thomas Leimk{\"u}hler, and George Drettakis.
\newblock 3d gaussian splatting for real-time radiance field rendering.
\newblock \emph{ACM Trans. Graph.}, 42\penalty0 (4):\penalty0 139--1, 2023.

\bibitem[Kingma(2013)]{kingma2013auto}
Diederik~P Kingma.
\newblock Auto-encoding variational bayes.
\newblock \emph{arXiv preprint arXiv:1312.6114}, 2013.

\bibitem[Lin et~al.(2023)Lin, Gao, Tang, Takikawa, Zeng, Huang, Kreis, Fidler, Liu, and Lin]{lin2023magic3d}
Chen-Hsuan Lin, Jun Gao, Luming Tang, Towaki Takikawa, Xiaohui Zeng, Xun Huang, Karsten Kreis, Sanja Fidler, Ming-Yu Liu, and Tsung-Yi Lin.
\newblock Magic3d: High-resolution text-to-3d content creation.
\newblock In \emph{Proceedings of the IEEE/CVF Conference on Computer Vision and Pattern Recognition}, pages 300--309, 2023.

\bibitem[Liu et~al.(2022)Liu, Gong, and Liu]{liu2022flow}
Xingchao Liu, Chengyue Gong, and Qiang Liu.
\newblock Flow straight and fast: Learning to generate and transfer data with rectified flow.
\newblock \emph{arXiv preprint arXiv:2209.03003}, 2022.

\bibitem[Liu et~al.(2024)Liu, Guo, Luo, Sun, Yin, and Zhang]{liu2024pi3d}
Ying-Tian Liu, Yuan-Chen Guo, Guan Luo, Heyi Sun, Wei Yin, and Song-Hai Zhang.
\newblock Pi3d: Efficient text-to-3d generation with pseudo-image diffusion.
\newblock In \emph{Proceedings of the IEEE/CVF Conference on Computer Vision and Pattern Recognition}, pages 19915--19924, 2024.

\bibitem[Lorensen and Cline(1998)]{lorensen1998marching}
William~E Lorensen and Harvey~E Cline.
\newblock Marching cubes: A high resolution 3d surface construction algorithm.
\newblock In \emph{Seminal graphics: pioneering efforts that shaped the field}, pages 347--353. 1998.

\bibitem[Lu et~al.(2024)Lu, Wang, Huang, Wu, Liu, Ouyang, and Bai]{lu2024fit}
Zeyu Lu, Zidong Wang, Di Huang, Chengyue Wu, Xihui Liu, Wanli Ouyang, and Lei Bai.
\newblock Fit: Flexible vision transformer for diffusion model.
\newblock \emph{arXiv preprint arXiv:2402.12376}, 2024.

\bibitem[Lugmayr et~al.(2022)Lugmayr, Danelljan, Romero, Yu, Timofte, and Van~Gool]{lugmayr2022repaint}
Andreas Lugmayr, Martin Danelljan, Andres Romero, Fisher Yu, Radu Timofte, and Luc Van~Gool.
\newblock Repaint: Inpainting using denoising diffusion probabilistic models.
\newblock In \emph{Proceedings of the IEEE/CVF conference on computer vision and pattern recognition}, pages 11461--11471, 2022.

\bibitem[Ma et~al.(2024)Ma, Goldstein, Albergo, Boffi, Vanden-Eijnden, and Xie]{ma2024sit}
Nanye Ma, Mark Goldstein, Michael~S Albergo, Nicholas~M Boffi, Eric Vanden-Eijnden, and Saining Xie.
\newblock Sit: Exploring flow and diffusion-based generative models with scalable interpolant transformers.
\newblock \emph{arXiv preprint arXiv:2401.08740}, 2024.

\bibitem[Mildenhall et~al.(2021)Mildenhall, Srinivasan, Tancik, Barron, Ramamoorthi, and Ng]{mildenhall2021nerf}
Ben Mildenhall, Pratul~P Srinivasan, Matthew Tancik, Jonathan~T Barron, Ravi Ramamoorthi, and Ren Ng.
\newblock Nerf: Representing scenes as neural radiance fields for view synthesis.
\newblock \emph{Communications of the ACM}, 65\penalty0 (1):\penalty0 99--106, 2021.

\bibitem[Nash et~al.(2020)Nash, Ganin, Eslami, and Battaglia]{nash2020polygen}
Charlie Nash, Yaroslav Ganin, SM~Ali Eslami, and Peter Battaglia.
\newblock Polygen: An autoregressive generative model of 3d meshes.
\newblock In \emph{International conference on machine learning}, pages 7220--7229. PMLR, 2020.

\bibitem[Oquab et~al.(2023)Oquab, Darcet, Moutakanni, Vo, Szafraniec, Khalidov, Fernandez, Haziza, Massa, El-Nouby, et~al.]{oquab2023dinov2}
Maxime Oquab, Timoth{\'e}e Darcet, Th{\'e}o Moutakanni, Huy Vo, Marc Szafraniec, Vasil Khalidov, Pierre Fernandez, Daniel Haziza, Francisco Massa, Alaaeldin El-Nouby, et~al.
\newblock Dinov2: Learning robust visual features without supervision.
\newblock \emph{arXiv preprint arXiv:2304.07193}, 2023.

\bibitem[Park et~al.(2019)Park, Florence, Straub, Newcombe, and Lovegrove]{park2019deepsdf}
Jeong~Joon Park, Peter Florence, Julian Straub, Richard Newcombe, and Steven Lovegrove.
\newblock Deepsdf: Learning continuous signed distance functions for shape representation.
\newblock In \emph{Proceedings of the IEEE/CVF conference on computer vision and pattern recognition}, pages 165--174, 2019.

\bibitem[Peebles and Xie(2023)]{peebles2023scalable}
William Peebles and Saining Xie.
\newblock Scalable diffusion models with transformers.
\newblock In \emph{Proceedings of the IEEE/CVF International Conference on Computer Vision}, pages 4195--4205, 2023.

\bibitem[Poole et~al.(2022)Poole, Jain, Barron, and Mildenhall]{poole2022dreamfusion}
Ben Poole, Ajay Jain, Jonathan~T Barron, and Ben Mildenhall.
\newblock Dreamfusion: Text-to-3d using 2d diffusion.
\newblock \emph{arXiv preprint arXiv:2209.14988}, 2022.

\bibitem[Rombach et~al.(2022)Rombach, Blattmann, Lorenz, Esser, and Ommer]{rombach2022high}
Robin Rombach, Andreas Blattmann, Dominik Lorenz, Patrick Esser, and Bj{\"o}rn Ommer.
\newblock High-resolution image synthesis with latent diffusion models.
\newblock In \emph{Proceedings of the IEEE/CVF conference on computer vision and pattern recognition}, pages 10684--10695, 2022.

\bibitem[Siddiqui et~al.(2024)Siddiqui, Alliegro, Artemov, Tommasi, Sirigatti, Rosov, Dai, and Nie{\ss}ner]{siddiqui2024meshgpt}
Yawar Siddiqui, Antonio Alliegro, Alexey Artemov, Tatiana Tommasi, Daniele Sirigatti, Vladislav Rosov, Angela Dai, and Matthias Nie{\ss}ner.
\newblock Meshgpt: Generating triangle meshes with decoder-only transformers.
\newblock In \emph{Proceedings of the IEEE/CVF Conference on Computer Vision and Pattern Recognition}, pages 19615--19625, 2024.

\bibitem[Song et~al.(2020)Song, Sohl-Dickstein, Kingma, Kumar, Ermon, and Poole]{song2020score}
Yang Song, Jascha Sohl-Dickstein, Diederik~P Kingma, Abhishek Kumar, Stefano Ermon, and Ben Poole.
\newblock Score-based generative modeling through stochastic differential equations.
\newblock \emph{arXiv preprint arXiv:2011.13456}, 2020.

\bibitem[Su et~al.(2024)Su, Ahmed, Lu, Pan, Bo, and Liu]{su2024roformer}
Jianlin Su, Murtadha Ahmed, Yu Lu, Shengfeng Pan, Wen Bo, and Yunfeng Liu.
\newblock Roformer: Enhanced transformer with rotary position embedding.
\newblock \emph{Neurocomputing}, 568:\penalty0 127063, 2024.

\bibitem[Tang et~al.(2025)Tang, Chen, Chen, Wang, Zeng, and Liu]{tang2025lgm}
Jiaxiang Tang, Zhaoxi Chen, Xiaokang Chen, Tengfei Wang, Gang Zeng, and Ziwei Liu.
\newblock Lgm: Large multi-view gaussian model for high-resolution 3d content creation.
\newblock In \emph{European Conference on Computer Vision}, pages 1--18. Springer, 2025.

\bibitem[Touvron et~al.(2023)Touvron, Lavril, Izacard, Martinet, Lachaux, Lacroix, Rozi{\`e}re, Goyal, Hambro, Azhar, et~al.]{touvron2023llama}
Hugo Touvron, Thibaut Lavril, Gautier Izacard, Xavier Martinet, Marie-Anne Lachaux, Timoth{\'e}e Lacroix, Baptiste Rozi{\`e}re, Naman Goyal, Eric Hambro, Faisal Azhar, et~al.
\newblock Llama: Open and efficient foundation language models.
\newblock \emph{arXiv preprint arXiv:2302.13971}, 2023.

\bibitem[Van Den~Oord et~al.(2017)Van Den~Oord, Vinyals, et~al.]{van2017neural}
Aaron Van Den~Oord, Oriol Vinyals, et~al.
\newblock Neural discrete representation learning.
\newblock \emph{Advances in neural information processing systems}, 30, 2017.

\bibitem[Weng et~al.(2024)Weng, Wang, Zhang, Chen, and Zhu]{weng2024pivotmesh}
Haohan Weng, Yikai Wang, Tong Zhang, CL Chen, and Jun Zhu.
\newblock Pivotmesh: Generic 3d mesh generation via pivot vertices guidance.
\newblock \emph{arXiv preprint arXiv:2405.16890}, 2024.

\bibitem[Xu et~al.(2024)Xu, Shi, Yifan, Chen, Yang, Peng, Shen, and Wetzstein]{xu2024grm}
Yinghao Xu, Zifan Shi, Wang Yifan, Hansheng Chen, Ceyuan Yang, Sida Peng, Yujun Shen, and Gordon Wetzstein.
\newblock Grm: Large gaussian reconstruction model for efficient 3d reconstruction and generation.
\newblock \emph{arXiv preprint arXiv:2403.14621}, 2024.

\bibitem[Zeghidour et~al.(2021)Zeghidour, Luebs, Omran, Skoglund, and Tagliasacchi]{zeghidour2021soundstream}
Neil Zeghidour, Alejandro Luebs, Ahmed Omran, Jan Skoglund, and Marco Tagliasacchi.
\newblock Soundstream: An end-to-end neural audio codec.
\newblock \emph{IEEE/ACM Transactions on Audio, Speech, and Language Processing}, 30:\penalty0 495--507, 2021.

\bibitem[Zhao et~al.(2024)Zhao, Liu, Chen, Zeng, Wang, Cheng, Fu, Chen, Yu, and Gao]{zhao2024michelangelo}
Zibo Zhao, Wen Liu, Xin Chen, Xianfang Zeng, Rui Wang, Pei Cheng, Bin Fu, Tao Chen, Gang Yu, and Shenghua Gao.
\newblock Michelangelo: Conditional 3d shape generation based on shape-image-text aligned latent representation.
\newblock \emph{Advances in Neural Information Processing Systems}, 36, 2024.

\bibitem[Zhuo et~al.(2024)Zhuo, Du, Xiao, Li, Liu, Huang, Liu, Zhao, Wang, Ma, et~al.]{zhuo2024lumina}
Le Zhuo, Ruoyi Du, Han Xiao, Yangguang Li, Dongyang Liu, Rongjie Huang, Wenze Liu, Lirui Zhao, Fu-Yun Wang, Zhanyu Ma, et~al.
\newblock Lumina-next: Making lumina-t2x stronger and faster with next-dit.
\newblock \emph{arXiv preprint arXiv:2406.18583}, 2024.

\end{thebibliography}
